\documentclass{article} 
\usepackage{iclr2026_conference,times}
\iclrfinalcopy

\usepackage{amsmath,amsfonts,bm}

\def\eqref#1{equation~\ref{#1}}

\def\1{\bm{1}}

\DeclareMathAlphabet{\mathsfit}{\encodingdefault}{\sfdefault}{m}{sl}
\SetMathAlphabet{\mathsfit}{bold}{\encodingdefault}{\sfdefault}{bx}{n}

\usepackage[utf8]{inputenc} 
\usepackage[T1]{fontenc}    
\usepackage{hyperref}       
\usepackage{url}            
\usepackage{booktabs}       
\usepackage{amsfonts}       
\usepackage{nicefrac}       
\usepackage{microtype}      
\usepackage{xcolor}         
\usepackage{graphicx}
\usepackage{subcaption}
\usepackage{wrapfig}
\usepackage{algorithm}
\usepackage{algorithmic}
\usepackage{caption}
\usepackage{amsmath}
\let\cite\citep

\title{Verifier Threshold: An Efficient Test-Time Scaling Approach for Image Generation\\[0.1em]}

\author{
\renewcommand{\arraystretch}{1.1}
\begin{tabular}{@{}c@{\hspace{2.5em}}c@{}}
\begin{tabular}[t]{@{}c@{}}
\textbf{Vignesh Sundaresha} \\
University of Illinois Urbana-Champaign \\
\texttt{vs49@illinois.edu}
\end{tabular}
&
\begin{tabular}[t]{@{}c@{}}
\textbf{Akash Haridas} \\
AMD \\
\texttt{akash.haridas@amd.com}
\end{tabular}
\\
\multicolumn{2}{c}{\rule{0pt}{1.4em}} \\
\begin{tabular}[t]{@{}c@{}}
\textbf{Vikram Appia} \\
AMD \\
\end{tabular}
&
\begin{tabular}[t]{@{}c@{}}
\textbf{Lav R.\ Varshney} \\
Stony Brook University \\
\end{tabular}
\end{tabular}
}

\begin{document}

\maketitle
\vspace{-15pt}
\begin{abstract}
\vspace{-10pt}
Image generation has emerged as a mainstream application of large generative models. Just as test-time compute and reasoning have improved language model capabilities, similar benefits have been observed for image generation models. In particular, searching over noise samples for diffusion and flow models has been shown to scale well with test-time compute. While recent works explore allocating non-uniform inference-compute budgets across denoising steps, existing approaches rely on greedy heuristics and often allocate the compute budget ineffectively. In this work, we study this problem and propose a simple fix. We propose \textit{Verifier-Threshold}, which automatically reallocates test-time compute and delivers substantial efficiency improvements. For the same performance on the GenEval benchmark, we achieve a $2\times$-$\;4\times$ reduction in computational time over the state-of-the-art method.
\end{abstract}

\vspace{-10pt}
\section{Introduction}  
\vspace{-5pt}
\begin{wrapfigure}{r}{0.45\linewidth}
    \vspace{-10pt}
    \centering
    \vspace{-1.5em} 
    \includegraphics[width=\linewidth]{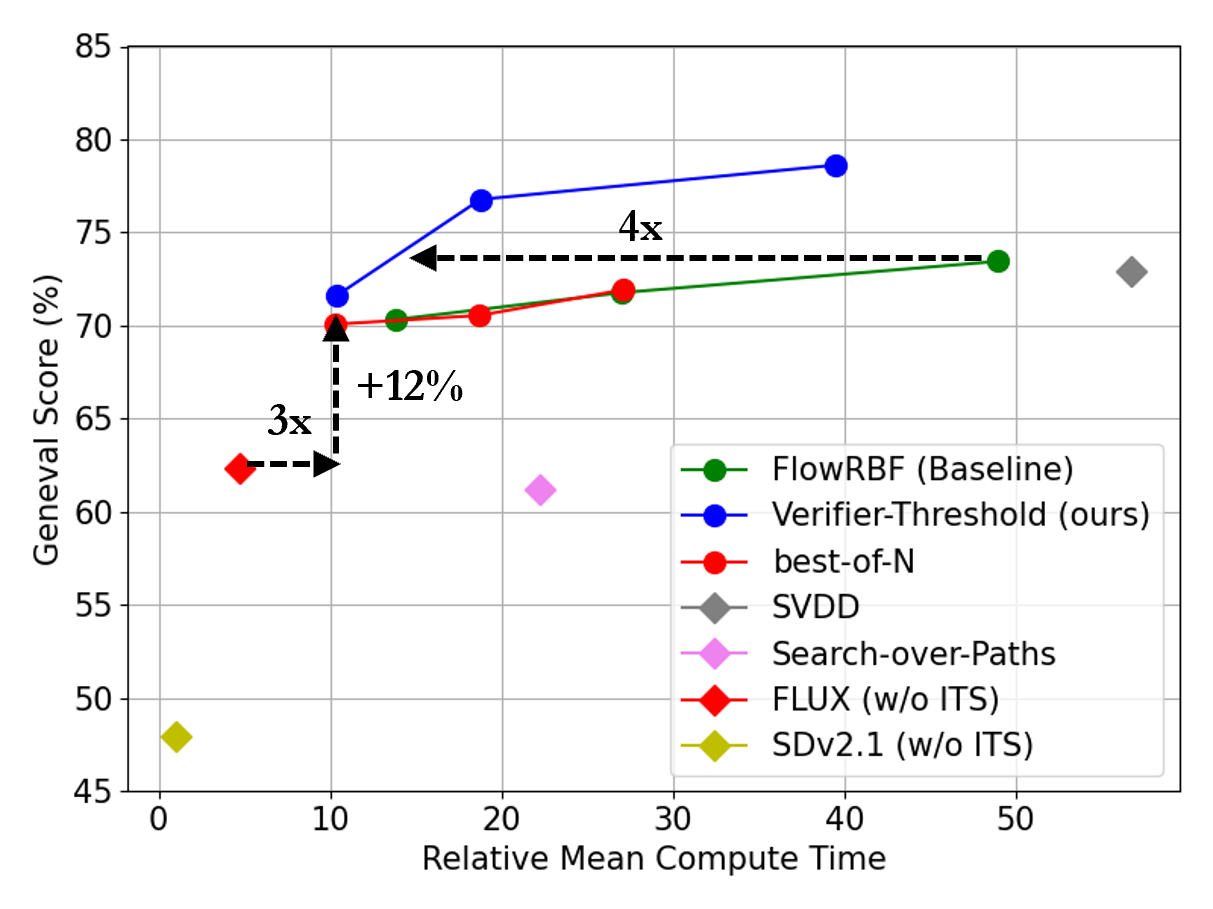}
    \caption{Test-time scaling; the horizontal axis represents relative wall-clock duration on an AMD MI300X GPU averaged across 553 GenEval~\cite{ghosh2023geneval} prompts and the vertical axis represents the overall GenEval score. VQAScore~\cite{lin2024evaluating} is used as the verifier and FLUX-Schnell~\cite{labs2025flux1kontextflowmatching} as the generator for all methods. \textit{Verifier-Threshold} is consistently more efficient and delivers higher scores than the baseline FlowRBF and other methods.}
    \label{fig: intro}
    \vspace{-10pt}
\end{wrapfigure}

Image generation has become a popular application of AI, with several models~\cite{google2025nanoBanana, openai2025_gptimage1_api, apple2025_univg, midjourney2025_v7, stability2024_sd35} now widely used. However, even state-of-the-art models often struggle to accurately follow detailed prompt instructions~\cite{ghosh2023geneval, huang2023t2i}. To address this limitation, the vision community has drawn inspiration from language models by incorporating test-time compute~\cite{snell2024scaling} and reasoning~\cite{guo2025deepseek}, leading to the development of analogous approaches for vision tasks~\cite{ma2025inference, zhuo2025reflection}. Reasoning-based methods, however, inherently demand substantial computation and wall-clock time at inference, which becomes costly at scale~\cite{jin2025energy} or on resource-constrained devices~\cite{stelia2025_reasoningEdge}. Consequently, improving the efficiency of test-time compute algorithms is essential.

Fig.~\ref{fig: intro} illustrates how test-time scaling increases computational requirements. The baseline implementation~\cite{kim2025inference} requires between $3\times$ and $10\times$ more compute time than regular image generation to achieve GenEval~\cite{ghosh2023geneval} gains of $8\%$ to $12\%$, respectively. In contrast, our proposed method, \textit{Verifier-Threshold}, reaches the same $12\%$ improvement using only $(1/4)^{\text{th}}$ of the baseline's computational cost, clearly demonstrating its efficiency. Furthermore, our algorithm continues to benefit from scaling, achieving gains of up to $15\%$.

The key contributions of this work are twofold: \textit{identifying the ``compute-dumping'' issue} in an existing baseline algorithm and \textit{introducing the \textit{Verifier-Threshold} algorithm} as a solution. We validate our approach on state-of-the-art image generation models and datasets. We organize the rest of the paper as follows: Sec.~\ref{sec: prelimnaries} discusses the baseline algorithm and experimental setup. Sec.~\ref{sec: VT} analyzes the compute dumping phenomenon and presents solutions, including \textit{Verifier-Threshold}. Lastly, Sec.~\ref{sec: results} presents results, including example images generated by the different approaches.

\vspace{-10pt}
\section{Preliminaries}
\vspace{-7pt}
\label{sec: prelimnaries}
\subsection{Related Works}
\vspace{-5pt}
Diffusion models~\cite{dhariwal2021diffusion, ho2020denoising} and flow models~\cite{labs2025flux1kontextflowmatching} along with autoregressive models~\cite{tian2024visual} have become the mainstream methods for image generation. Test-time scaling for image generation basically aims to use the compute available at test-time to improve performance, either through increasing the denoising steps~\cite{salimans2022progressive} or searching for better noises~\cite{ma2025inference, kim2025inference}.
Noise-search in particular has been shown to scale much better than increasing denoising steps~\cite{ma2025inference}; hence we focus on it here. FlowRBF introduces the key idea of \textit{non-uniform inference-compute budget allocation across denoising steps}, which is a crucial advance in noise-search approaches. Using this idea, it outperforms prior baselines such as Best-of-$N$ (which performs as well as zero-order and Search-over-Paths) from~\cite{ma2025inference}. Hence, we use FlowRBF as our main baseline and show improvements on the same flow models as a proof of concept. FlowRBF's approach of non-uniform allocation and our proposed algorithm can also be extended to other diffusion models. Prior work is discussed in detail in Appendix~\ref{app: related work}.

\vspace{-5pt}
\subsection{Experimental Setup}
\vspace{-5pt}
We use the FLUX-Schnell model~\cite{labs2025flux1kontextflowmatching} for all experiments, following~\cite{kim2025inference}. We use ImageReward~\cite{xu2023imagereward} and VQAScore~\cite{lin2024evaluating} as verifier reward models to guide the noise-search process. Only open-source models are used throughout since these experiments cannot be conducted with closed-source models. To evaluate performance, we use the GenEval benchmark~\cite{ghosh2023geneval}, which consists of prompts that test for attributes such as color, number, and position and provides objective answers. We report the average percentage of correct images across attributes as our performance metric. For efficiency, we use the average wall-clock duration per prompt on a single AMD MI300X GPU (reported relative to SDv2.1 which takes the least amount of time; see Appendix~\ref{app: results}). We also report the number of function evaluations (NFEs), where lower is better. More details are provided in Appendix~\ref{app: experimental setup}.

\vspace{-5pt}
\section{Verifier Threshold}
\vspace{-5pt}
\label{sec: VT}

\subsection{Compute Dumping and Manual Budgeting}

\begin{figure}[th]
    \centering
    \begin{subfigure}{0.24\linewidth}
        \centering
        \includegraphics[width=\linewidth]{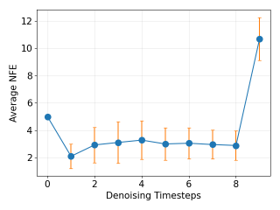}
        \caption{}
    \end{subfigure}
    \hfill
    \begin{subfigure}{0.24\linewidth}
        \centering
        \includegraphics[width=\linewidth]{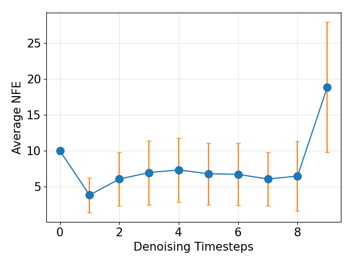}
        \caption{}
    \end{subfigure}
    \hfill
    \begin{subfigure}{0.24\linewidth}
        \centering
        \includegraphics[width=\linewidth]{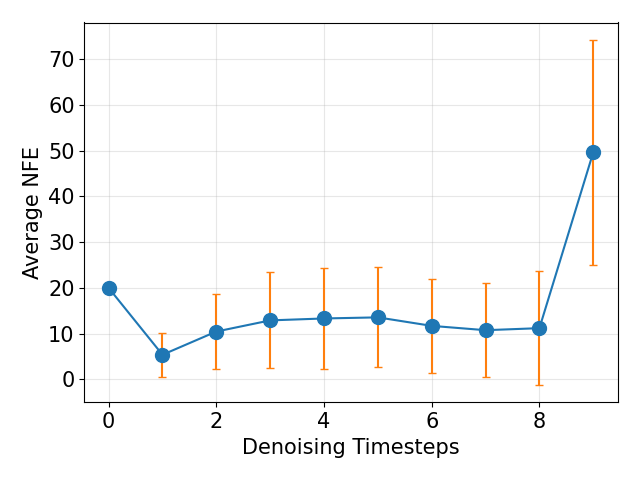}
        \caption{}
    \end{subfigure}
    \hfill
    \begin{subfigure}{0.24\linewidth}
        \centering
        \includegraphics[width=\linewidth]{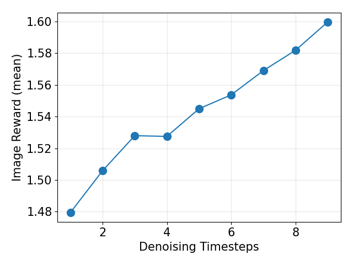}
        \caption{}
    \end{subfigure}

    \caption{(a), (b) and (c) show how FlowRBF distributes NFEs (or compute budget) across denoising steps, averaged across all GenEval prompts with the total NFE budget set to 40, 80, and 160, respectively. Notice that in all cases, the inference-compute gets "dumped" at the last denoising step, and this issue persists even at total NFE budget as large as 500. (d) shows the verifier score as a function of denoising steps averaged over all GenEval prompts for total NFEs = 80. Notice that the verifier score does not increase at the last step in proportion to the amount of compute dedicated to it, thus wasting the test-time compute budget.}
    \label{fig: compute dumping}
\end{figure}

The efficacy of FlowRBF can be mainly attributed to its non-uniform inference-compute allocation across denoising steps, termed \emph{rollover budget forcing (RBF)}.
The basic concept behind RBF is that the denoiser remains at a denoising step until either: (a) the verifier score for a new noise particle exceeds the previous step's best score, or (b) the per-step maximum compute budget is exhausted. The issue with these criteria is that (a) the method is greedy and advances even for marginal improvements, and (b) it does not account for ``wrong'' trajectories induced by per-step budget constraints.

\begin{figure}[h]
    \centering
    \begin{minipage}{0.72\linewidth}
        \centering
        \includegraphics[width=0.32\linewidth]{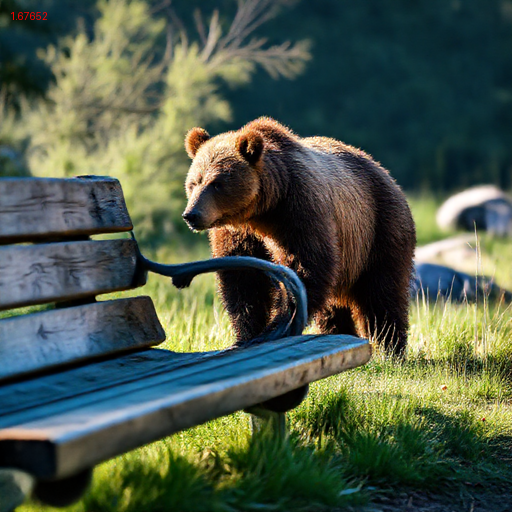}
        \includegraphics[width=0.32\linewidth]{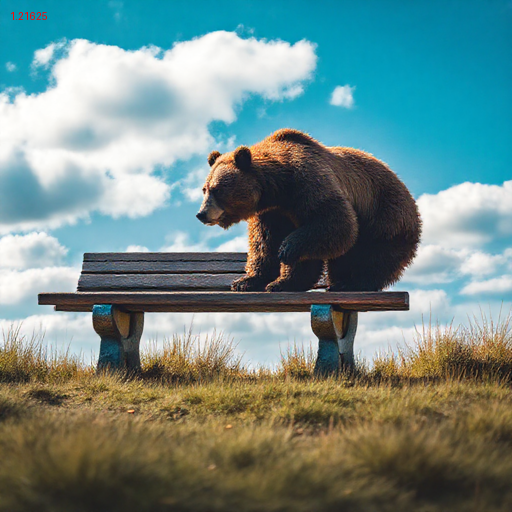}
        \includegraphics[width=0.32\linewidth]{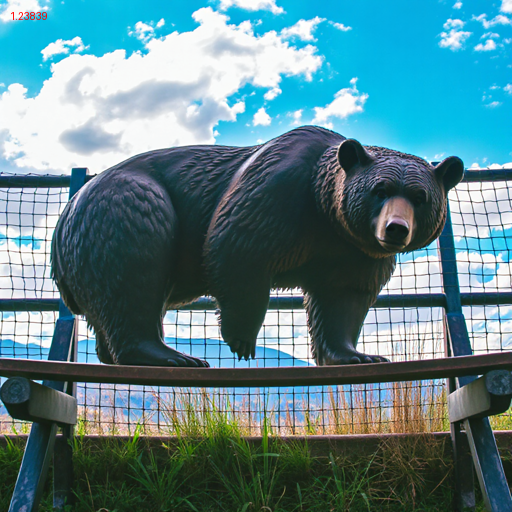}
        
        \subcaption*{(a)} 
    \end{minipage}
    \hfill
    \begin{minipage}{0.25\linewidth}
        \centering
        \includegraphics[width=\linewidth]{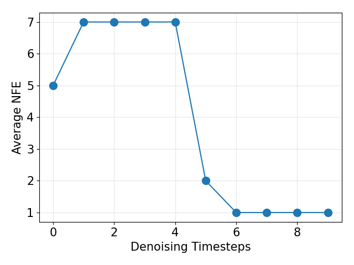}
        
        \subcaption*{(b)}
    \end{minipage}

    \caption{(a) shows three images generated at different denoising steps for the prompt “a photo of a bench left of a bear” using FlowRBF, with a total of 40 NFEs and ImageReward~\cite{xu2023imagereward} as the verifier. This example illustrates the compute-dumping issue. The first image (at the second denoising step) captures the correct structure, though the texture remains noisy. However, the second image, produced at the third step, achieves the highest verifier score and is therefore selected, despite placing the bear on top of the bench rather than to the left. Because no additional budget is allocated at earlier steps, the denoising process continues from this incorrect latent, ultimately producing a final image with the bear on top of the bench. (b) demonstrates how this issue could be mitigated by manually reallocating the compute budget through hard-coded NFEs at each step (i.e., manual budgeting).}
    \label{fig: mistake and manual budgetting}
\end{figure}

Manifestations of these issues are shown in Figs.~\ref{fig: compute dumping} and~\ref{fig: mistake and manual budgetting}, where the model runs a ten-step denoising process. Fig.~\ref{fig: compute dumping}(a--c) show that although RBF is designed to enable non-uniform allocation of inference compute (NFEs), most steps (1--8) receive similar NFEs (except for step 0, which is hard-coded to use a predefined NFE). As a result, a large fraction of the test-time NFEs are ``dumped'' into the final step. When the NFE budget is lower (\(=40\); Fig.~\ref{fig: compute dumping}(a)), the algorithm quickly exhausts the per-step budget and advances to the next step. When the budget is higher (\(=80,160\); Fig.~\ref{fig: compute dumping}(b,c)), the greediness of RBF causes it to advance quickly to the next step as well. In both cases, substantial budget remains and is forced into the last denoising step. Fig.~\ref{fig: compute dumping}(d) shows that the average verifier score at the last denoising step does not increase proportionally with the allocated compute, underscoring the dumping issue. Fig.~\ref{fig: mistake and manual budgetting}(a) shows an example prompt that highlights the resulting failure mode.

Intuitively, the RBF budget allocations in Fig.~\ref{fig: compute dumping} are sub-optimal because the structural changes needed to obtain a semantically correct image often occur in early denoising steps, suggesting that more budget should be allocated early rather than at the end. Manually hard-coding the compute distribution to allocate more budget to initial steps is one way to address this issue (Fig.~\ref{fig: mistake and manual budgetting}(b)). Although this can improve some prompts, it does not generalize well. Therefore, an algorithm that automatically redistributes compute is needed.

\vspace{-10pt}

\subsection{Verifier-Threshold Algorithm}
\label{subsec: VT}
\begin{figure}[t]
  \centering
  \begin{minipage}[t]{0.55\linewidth}\vspace{0pt}
  \begin{algorithm}[H]
  \caption{Verifier-Threshold Algorithm}
      \label{alg: vt}
    \begin{algorithmic}[1]
      \STATE \textbf{Input}: compute budget $N$, denoising timesteps $T$, generator model $\mathcal{G}_{\theta}$, verifier model $\mathcal{R}$, prompt $p$, Verifier-Threshold $\Delta$
      \STATE \textbf{Output}: image
      \STATE $t=1$
      \FOR{$n = {1, ... N}$}  
      \STATE \COMMENT{sample a noise particle and perform 1 NFE}
      \STATE $\epsilon_n \sim \mathcal{N}(\mathbf{0, I})$
        \STATE $z_t = \mathcal{G}_{\theta}(z_{t-1}, \epsilon_n, t, p)$
        \STATE \COMMENT{calculate reward score using verifier}
        \STATE $r_t = \mathcal{R}(\text{Tweedie's}(z_t), p)$
        \STATE \COMMENT{Verifier-Threshold criteria}
        \IF{$r_t - r_{t-1} > \Delta$ :}
        \STATE $t++$
        \ENDIF
        \IF{$t > T$ :}
        \STATE break
        \ENDIF
      \ENDFOR
      \RETURN Tweedie's ($z_t$)
    \end{algorithmic}
    \end{algorithm}
  \end{minipage}
  \hfill
    \begin{minipage}[t]{0.3\linewidth}\vspace{0pt}
    \centering
    \begin{subfigure}{\linewidth}
      \centering
      \includegraphics[width=\linewidth]{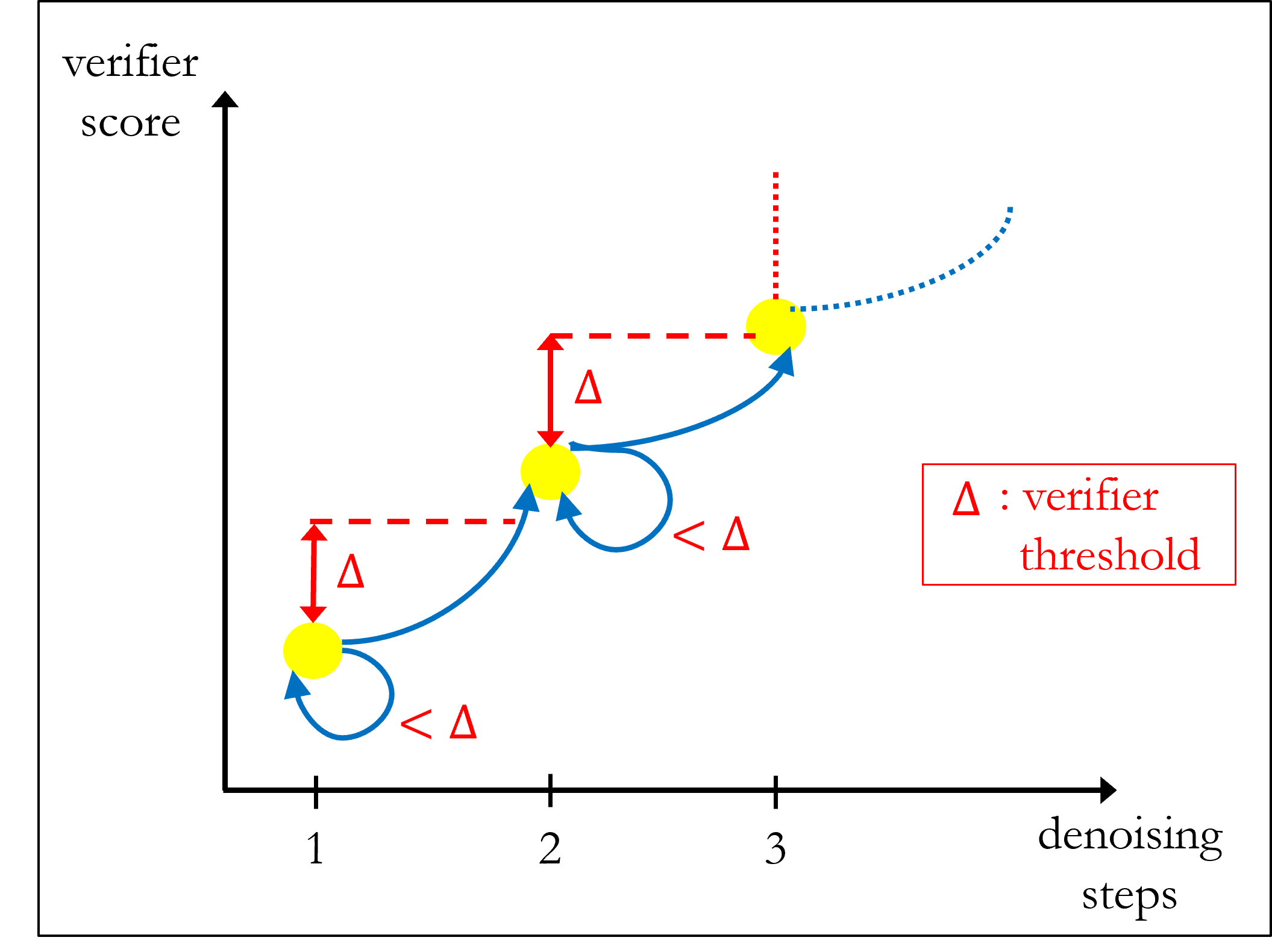}
      \subcaption{}
      \label{fig: vt}
    \end{subfigure}


    \begin{subfigure}{\linewidth}
      \centering
      \includegraphics[width=\linewidth]{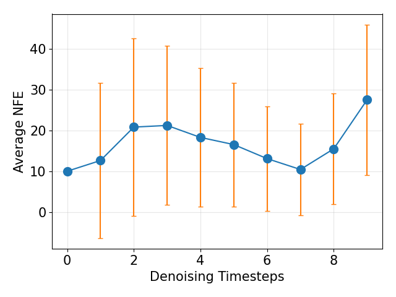}
      \subcaption{}
      \label{fig: vt budget allocation}
    \end{subfigure}

    \captionof{figure}{(a) verifier threshold mechanism and 
    (b) its compute allocation.}
    \label{fig:vt}
  \end{minipage}
\end{figure}

\vspace{-5pt}
Rather than advancing to the next denoising step as soon as a higher-reward noise particle is obtained, we propose waiting until the verifier score surpasses the previous one by a significant margin (i.e., a threshold). In this way, the algorithm progresses only when the score at a given step improves sufficiently. This mechanism forms the basis of our \textit{Verifier-Threshold} algorithm, as illustrated in Fig.~\ref{fig:vt}(a) and detailed in Alg.~\ref{alg: vt}.

\vspace{-10pt}
\section{Results}
\label{sec: results}
\vspace{-10pt}

\begin{wrapfigure}{r}{0.4\linewidth}
    \centering
    \vspace{-40pt}
    \vspace{-1em} 
    \includegraphics[width=\linewidth]{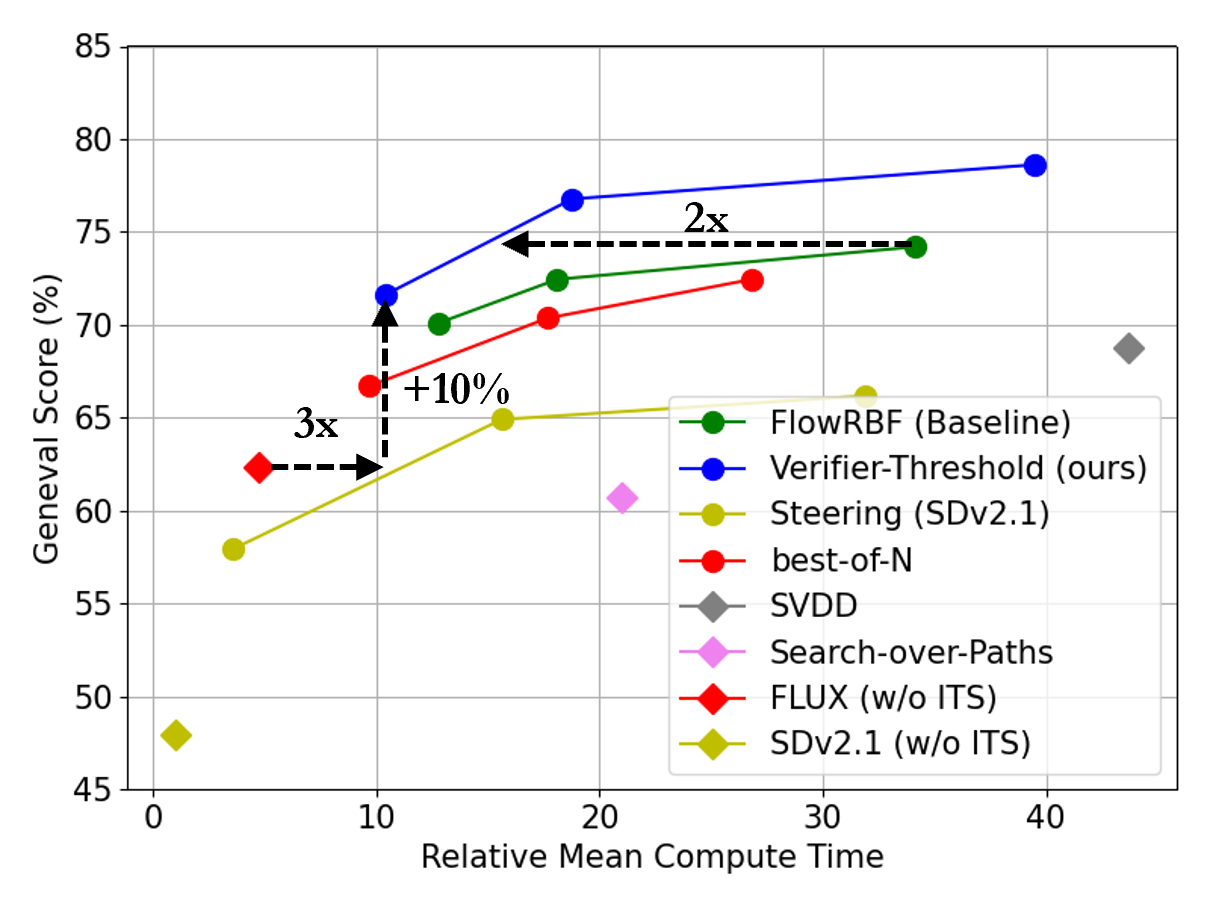}
    \caption{Test-time scaling when ImageReward~\cite{xu2023imagereward} is used as verifier.
    }
    \label{fig: results image reward}
\end{wrapfigure}

\begin{figure}[th]
    \centering
    \begin{subfigure}{0.30\linewidth}
        \centering
        \includegraphics[width=\linewidth]{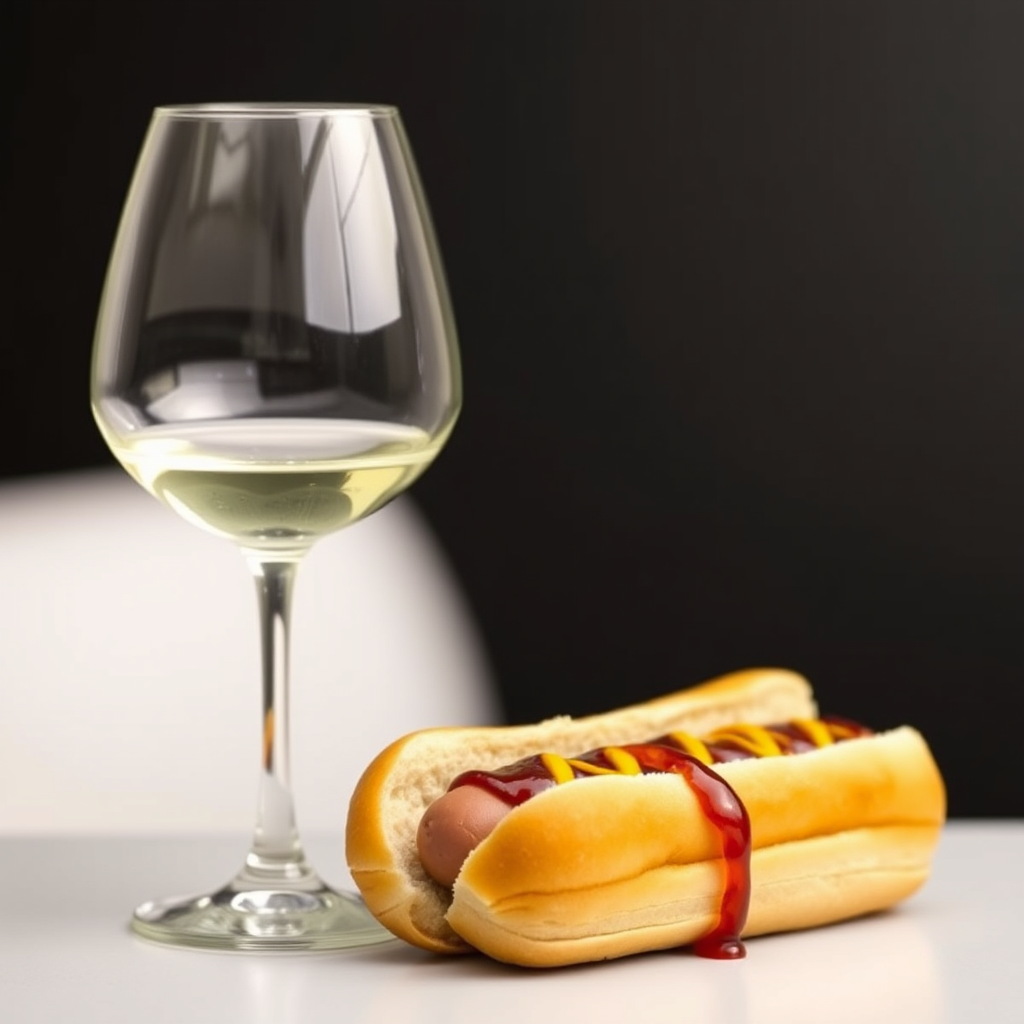}
        \caption{}
    \end{subfigure}
    \hfill
    \begin{subfigure}{0.30\linewidth}
        \centering
        \includegraphics[width=\linewidth]{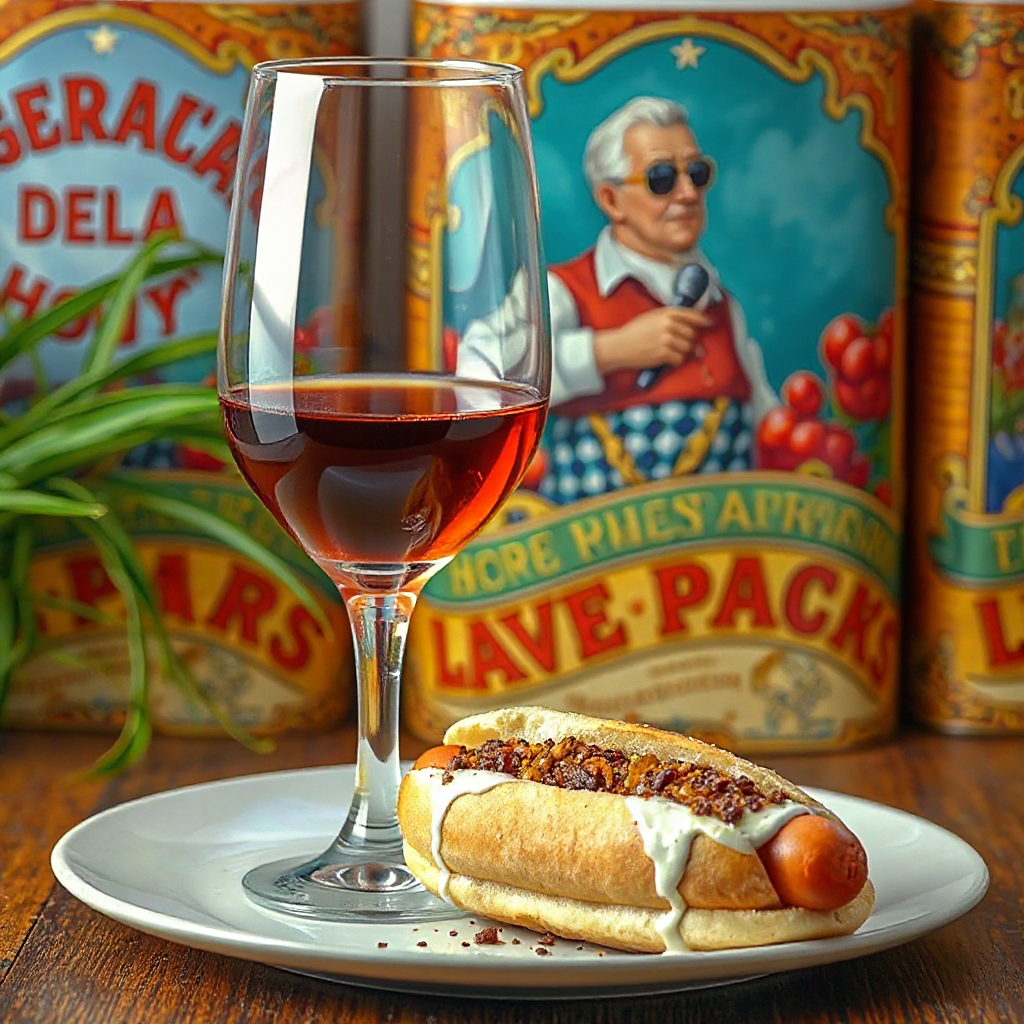}
        \caption{}
    \end{subfigure}
    \hfill
    \begin{subfigure}{0.30\linewidth}
        \centering
        \includegraphics[width=\linewidth]{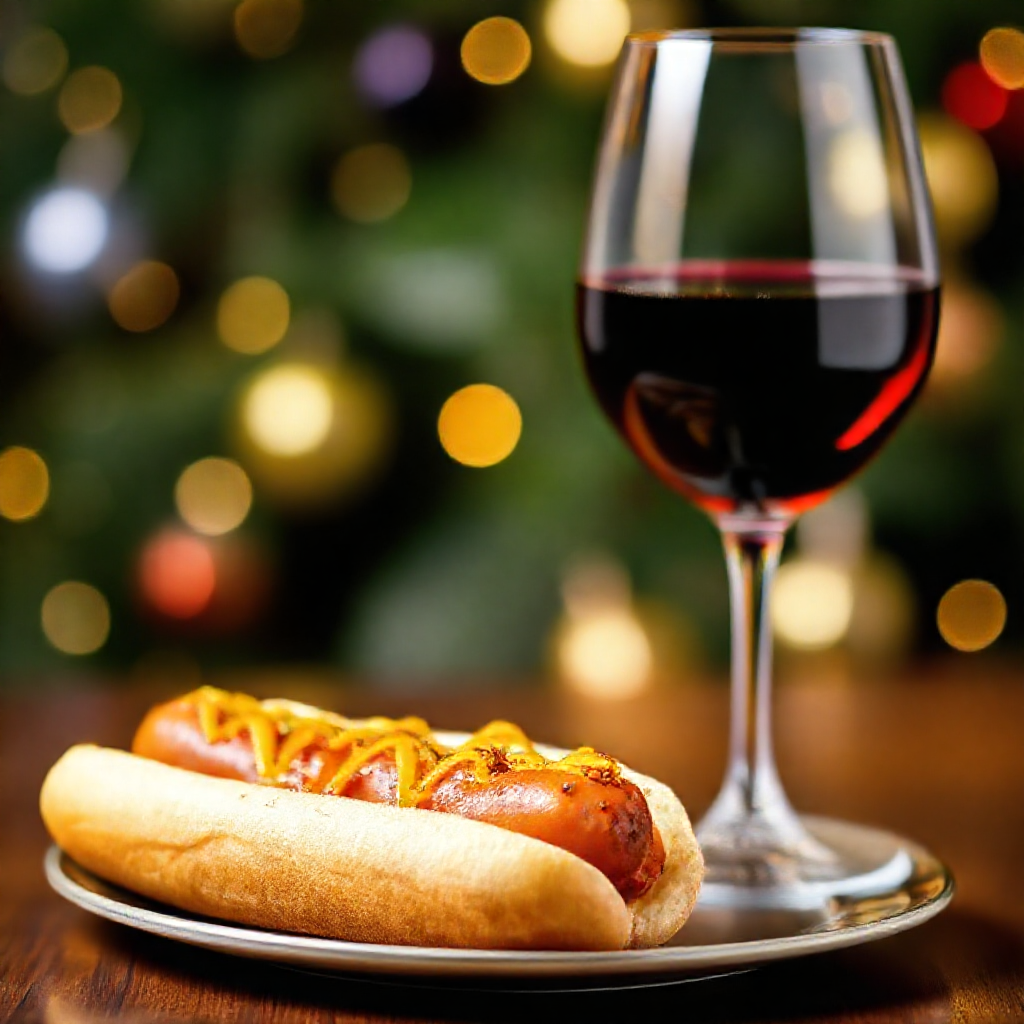}
        \caption{}
    \end{subfigure}

    \caption{The outputs of the model for the prompt ``a photo of a wine glass right of a hot dog''. (a), (b), (c) represent the images generated by FLUX-Schnell~\cite{labs2025flux1kontextflowmatching} with no test-time scaling, FlowRBF, and Verifier-Threshold, respectively. Figures (a) \& (b) have the wine glass on the left while (c) accurately provides what the prompt is requesting.}
    \label{fig: vt shining}
\end{figure}

As described in Sec.~\ref{sec: prelimnaries}, we evaluate different algorithms on the GenEval benchmark~\cite{ghosh2023geneval}. Fig.~\ref{fig: intro} highlights the advantages of our approach when using VQAScore~\cite{lin2024evaluating} as the verifier. The choice of threshold value \(\Delta\) is described in Appendix~\ref{app: experimental setup}. To show that these benefits generalize across verifiers, we also evaluate ImageReward~\cite{xu2023imagereward}, as shown in Fig.~\ref{fig: results image reward}. In both cases, our method achieves \(2\times\) to \(4\times\) higher efficiency for the same benchmark score, or alternatively a \(4\%\) to \(5\%\) improvement in benchmark score at the same computational cost. All reported results correspond to practical test-time compute budgets (2--3 minutes). Additionally, Fig.~\ref{fig: vt shining} presents example prompts where our algorithm produces objectively correct outputs compared to baseline methods. More comprehensive results and further examples are provided in Appendix~\ref{app: experimental setup}.

\vspace{-15pt}
\section{Conclusion}
\vspace{-10pt}
In this work, we identify a key flaw (compute dumping) in the state-of-the-art test-time scaling algorithm \cite{kim2025inference} and propose the Verifier-Threshold algorithm as an effective solution. We demonstrate its performance and efficiency advantages across multiple verifiers on the GenEval benchmark. For future work, we plan to investigate adaptive verifier thresholds that vary across denoising steps and explore strategies to automate their selection. There is also a lack of theoretical or interpretability grounding for test-time compute in the image generation community currently, and we will look into that as well as a future direction for our work.



\newpage
\onecolumn
\bibliography{iclr2026_conference}

@misc{google2025nanoBanana,
  title        = {Nano Banana: Gemini 2.5 Flash Image},
  author       = {Google DeepMind},
  year         = {2025},
  howpublished = {\\url{https://blog.google/products/gemini/updated-image-editing-model/}},
  note         = {Image generation / editing model, announcement August 26, 2025}
}

@misc{apple2025_univg,
  title        = {UniVG: A Generalist Diffusion Model for Unified Image Generation and Editing},
  author       = {{Apple Machine Learning Research}},
  year         = {2025},
  howpublished = {\url{https://machinelearning.apple.com/research/univg-diffusion-model}},
  note         = {Apple ML Research page describing the UniVG image generation/editing model}
}

@misc{midjourney2025_v7,
  title        = {Version 7},
  author       = {{Midjourney}},
  year         = {2025},
  month        = apr,
  howpublished = {\url{https://docs.midjourney.com/hc/en-us/articles/32199405667853-Version}},
  note         = {V7 released Apr 3, 2025; default since Jun 17, 2025}
}

@misc{stability2024_sd35,
  title        = {Introducing Stable Diffusion 3.5},
  author       = {{Stability AI}},
  year         = {2024},
  month        = oct,
  howpublished = {\url{https://stability.ai/news/introducing-stable-diffusion-3-5}},
  note         = {Launch post for SD~3.5 (Large / Large Turbo / Medium)}
}

@misc{openai2025_gptimage1_api,
  title        = {Introducing our latest image generation model in the API},
  author       = {{OpenAI}},
  year         = {2025},
  month        = apr,
  howpublished = {\url{https://openai.com/index/image-generation-api/}},
  note         = {API access to \texttt{gpt-image-1} (same image system as 4o in ChatGPT)}
}

@article{ghosh2023geneval,
  title={Geneval: An object-focused framework for evaluating text-to-image alignment},
  author={Ghosh, Dhruba and Hajishirzi, Hannaneh and Schmidt, Ludwig},
  journal={Advances in Neural Information Processing Systems},
  volume={36},
  pages={52132--52152},
  year={2023}
}

@article{huang2023t2i,
  title={T2i-compbench: A comprehensive benchmark for open-world compositional text-to-image generation},
  author={Huang, Kaiyi and Sun, Kaiyue and Xie, Enze and Li, Zhenguo and Liu, Xihui},
  journal={Advances in Neural Information Processing Systems},
  volume={36},
  pages={78723--78747},
  year={2023}
}

@article{snell2024scaling,
  title={Scaling llm test-time compute optimally can be more effective than scaling model parameters},
  author={Snell, Charlie and Lee, Jaehoon and Xu, Kelvin and Kumar, Aviral},
  journal={arXiv preprint arXiv:2408.03314},
  year={2024}
}

@article{guo2025deepseek,
  title={Deepseek-r1: Incentivizing reasoning capability in llms via reinforcement learning},
  author={Guo, Daya and Yang, Dejian and Zhang, Haowei and Song, Junxiao and Zhang, Ruoyu and Xu, Runxin and Zhu, Qihao and Ma, Shirong and Wang, Peiyi and Bi, Xiao and others},
  journal={arXiv preprint arXiv:2501.12948},
  year={2025}
}

@article{ma2025inference,
  title={Inference-time scaling for diffusion models beyond scaling denoising steps},
  author={Ma, Nanye and Tong, Shangyuan and Jia, Haolin and Hu, Hexiang and Su, Yu-Chuan and Zhang, Mingda and Yang, Xuan and Li, Yandong and Jaakkola, Tommi and Jia, Xuhui and others},
  journal={arXiv preprint arXiv:2501.09732},
  year={2025}
}

@article{zhuo2025reflection,
  title={From reflection to perfection: Scaling inference-time optimization for text-to-image diffusion models via reflection tuning},
  author={Zhuo, Le and Zhao, Liangbing and Paul, Sayak and Liao, Yue and Zhang, Renrui and Xin, Yi and Gao, Peng and Elhoseiny, Mohamed and Li, Hongsheng},
  journal={arXiv preprint arXiv:2504.16080},
  year={2025}
}

@article{jin2025energy,
  title={The Energy Cost of Reasoning: Analyzing Energy Usage in LLMs with Test-time Compute},
  author={Jin, Yunho and Wei, Gu-Yeon and Brooks, David},
  journal={arXiv preprint arXiv:2505.14733},
  year={2025}
}

@online{stelia2025_reasoningEdge,
  title        = {How Reasoning AI Models Are Transforming Edge Infrastructure},
  author       = {{Stelia}},
  year         = {2025},
  month        = mar,
  day          = {04},
  url          = {https://newsroom.stelia.ai/how-reasoning-ai-models-are-transforming-edge-infrastructure/},
  note         = {Accessed: YYYY-MM-DD}
}

@article{kim2025inference,
  title={Inference-time scaling for flow models via stochastic generation and rollover budget forcing},
  author={Kim, Jaihoon and Yoon, Taehoon and Hwang, Jisung and Sung, Minhyuk},
  journal={arXiv preprint arXiv:2503.19385},
  year={2025}
}

@article{dhariwal2021diffusion,
  title={Diffusion models beat gans on image synthesis},
  author={Dhariwal, Prafulla and Nichol, Alexander},
  journal={Advances in neural information processing systems},
  volume={34},
  pages={8780--8794},
  year={2021}
}

@article{ho2020denoising,
  title={Denoising diffusion probabilistic models},
  author={Ho, Jonathan and Jain, Ajay and Abbeel, Pieter},
  journal={Advances in neural information processing systems},
  volume={33},
  pages={6840--6851},
  year={2020}
}

@article{tian2024visual,
  title={Visual autoregressive modeling: Scalable image generation via next-scale prediction},
  author={Tian, Keyu and Jiang, Yi and Yuan, Zehuan and Peng, Bingyue and Wang, Liwei},
  journal={Advances in neural information processing systems},
  volume={37},
  pages={84839--84865},
  year={2024}
}

@misc{labs2025flux1kontextflowmatching,
      title={FLUX.1 Kontext: Flow Matching for In-Context Image Generation and Editing in Latent Space},
      author={Black Forest Labs and Stephen Batifol and Andreas Blattmann and Frederic Boesel and Saksham Consul and Cyril Diagne and Tim Dockhorn and Jack English and Zion English and Patrick Esser and Sumith Kulal and Kyle Lacey and Yam Levi and Cheng Li and Dominik Lorenz and Jonas Müller and Dustin Podell and Robin Rombach and Harry Saini and Axel Sauer and Luke Smith},
      year={2025},
      eprint={2506.15742},
      archivePrefix={arXiv},
      primaryClass={cs.GR},
      url={https://arxiv.org/abs/2506.15742},
}

@article{salimans2022progressive,
  title={Progressive distillation for fast sampling of diffusion models},
  author={Salimans, Tim and Ho, Jonathan},
  journal={arXiv preprint arXiv:2202.00512},
  year={2022}
}

@article{xu2023imagereward,
  title={Imagereward: Learning and evaluating human preferences for text-to-image generation},
  author={Xu, Jiazheng and Liu, Xiao and Wu, Yuchen and Tong, Yuxuan and Li, Qinkai and Ding, Ming and Tang, Jie and Dong, Yuxiao},
  journal={Advances in Neural Information Processing Systems},
  volume={36},
  pages={15903--15935},
  year={2023}
}

@inproceedings{lin2024evaluating,
  title={Evaluating text-to-visual generation with image-to-text generation},
  author={Lin, Zhiqiu and Pathak, Deepak and Li, Baiqi and Li, Jiayao and Xia, Xide and Neubig, Graham and Zhang, Pengchuan and Ramanan, Deva},
  booktitle={European Conference on Computer Vision},
  pages={366--384},
  year={2024},
  organization={Springer}
}

@article{singhal2025general,
  title={A general framework for inference-time scaling and steering of diffusion models},
  author={Singhal, Raghav and Horvitz, Zachary and Teehan, Ryan and Ren, Mengye and Yu, Zhou and McKeown, Kathleen and Ranganath, Rajesh},
  journal={arXiv preprint arXiv:2501.06848},
  year={2025}
}

@article{li2024derivative,
  title={Derivative-free guidance in continuous and discrete diffusion models with soft value-based decoding},
  author={Li, Xiner and Zhao, Yulai and Wang, Chenyu and Scalia, Gabriele and Eraslan, Gokcen and Nair, Surag and Biancalani, Tommaso and Ji, Shuiwang and Regev, Aviv and Levine, Sergey and others},
  journal={arXiv preprint arXiv:2408.08252},
  year={2024}
}

@inproceedings{radford2021learning,
  title={Learning transferable visual models from natural language supervision},
  author={Radford, Alec and Kim, Jong Wook and Hallacy, Chris and Ramesh, Aditya and Goh, Gabriel and Agarwal, Sandhini and Sastry, Girish and Askell, Amanda and Mishkin, Pamela and Clark, Jack and others},
  booktitle={International conference on machine learning},
  pages={8748--8763},
  year={2021},
  organization={PmLR}
}

@article{muennighoff2025s1,
  title={s1: Simple test-time scaling},
  author={Muennighoff, Niklas and Yang, Zitong and Shi, Weijia and Li, Xiang Lisa and Fei-Fei, Li and Hajishirzi, Hannaneh and Zettlemoyer, Luke and Liang, Percy and Cand{\`e}s, Emmanuel and Hashimoto, Tatsunori},
  journal={arXiv preprint arXiv:2501.19393},
  year={2025}
}

@misc{openai2024reason,
  author       = {OpenAI},
  title        = {Learning to Reason with LLMs},
  year         = {2024},
  howpublished = {\url{https://openai.com/index/learning-to-reason-with-llms/}},
  note         = {Accessed: 2025-09-29}
}

@article{svdd,
  title={Derivative-free guidance in continuous and discrete diffusion models with soft value-based decoding},
  author={Li, Xiner and Zhao, Yulai and Wang, Chenyu and Scalia, Gabriele and Eraslan, Gokcen and Nair, Surag and Biancalani, Tommaso and Ji, Shuiwang and Regev, Aviv and Levine, Sergey and others},
  journal={arXiv preprint arXiv:2408.08252},
  year={2024}
}
\bibliographystyle{iclr2026_conference}

\newpage
\appendix
\section{Experimental Setup}
\label{app: experimental setup}
\vspace{-5pt}
\subsection{Dataset \& Models}
\vspace{-5pt}
For evaluating our models we employ the widely-used GenEval dataset~\cite{ghosh2023geneval}. It contains 553 prompts which checks six different attributes: single object, two object, counting, colors, position, and attribute binding. For the generator, we use the FLUX-Schnell model~\cite{labs2025flux1kontextflowmatching}  extensively due to its SOTA performance and baseline implementation from FlowRBF. We run it consistently for ten denoising steps for best results. For the verifiers, we use ImageReward~\cite{xu2023imagereward} and VQAScore~\cite{lin2024evaluating} which are both variants of the CLIP model~\cite{radford2021learning}.

\vspace{-5pt}
\subsection{Hyperparameters \& Verifier-Threshold Choice}
\vspace{-5pt}
We use the same setup and hyperparameters as FlowRBF for all the methods: Best-of-$N$, search-over-paths, SVDD, FlowRBF and ours. When using 40 NFEs, we used 5 initial noise samples, and doubled/quadrupled it for 80/160 respectively. We used guidance-scale 3.5, diffusion norm 3.0, exponential time-scheduler with $t_{\text{max}}=1000$ and image-size 1024 $\times$ 1024.
\begin{figure}[H]
\centering
\begin{minipage}{0.62\linewidth}
\setlength{\parindent}{0pt}
The threshold value $\Delta$ is chosen independently for each verifier. The difference between the average verifier score for the initial timestep and final timestep is divided by the number of timesteps to obtain it. We use a verifier threshold value ($\Delta$) of 0.005 and 0.00125 for ImageReward and VQAScore respectively when the total NFEs is 40. For subsequent total NFEs of 80 and 160, we correspondingly multiplied the threshold values with $2\times, 4\times$ the above value. Fig.~\ref{fig: ablation_vt} shows the verifier score variation for ImageReward which was obtained when running a simple evaluation of the FlowRBF algorithm (NFE$=80$) for GenEval prompts. From the explanation above, we can calculate the value as $\Delta = \frac{1.62-1.52}{10} = 0.01 = 2\times 0.005$.

\end{minipage}\hfill
\begin{minipage}{0.35\linewidth}
\centering
\includegraphics[width=\linewidth]{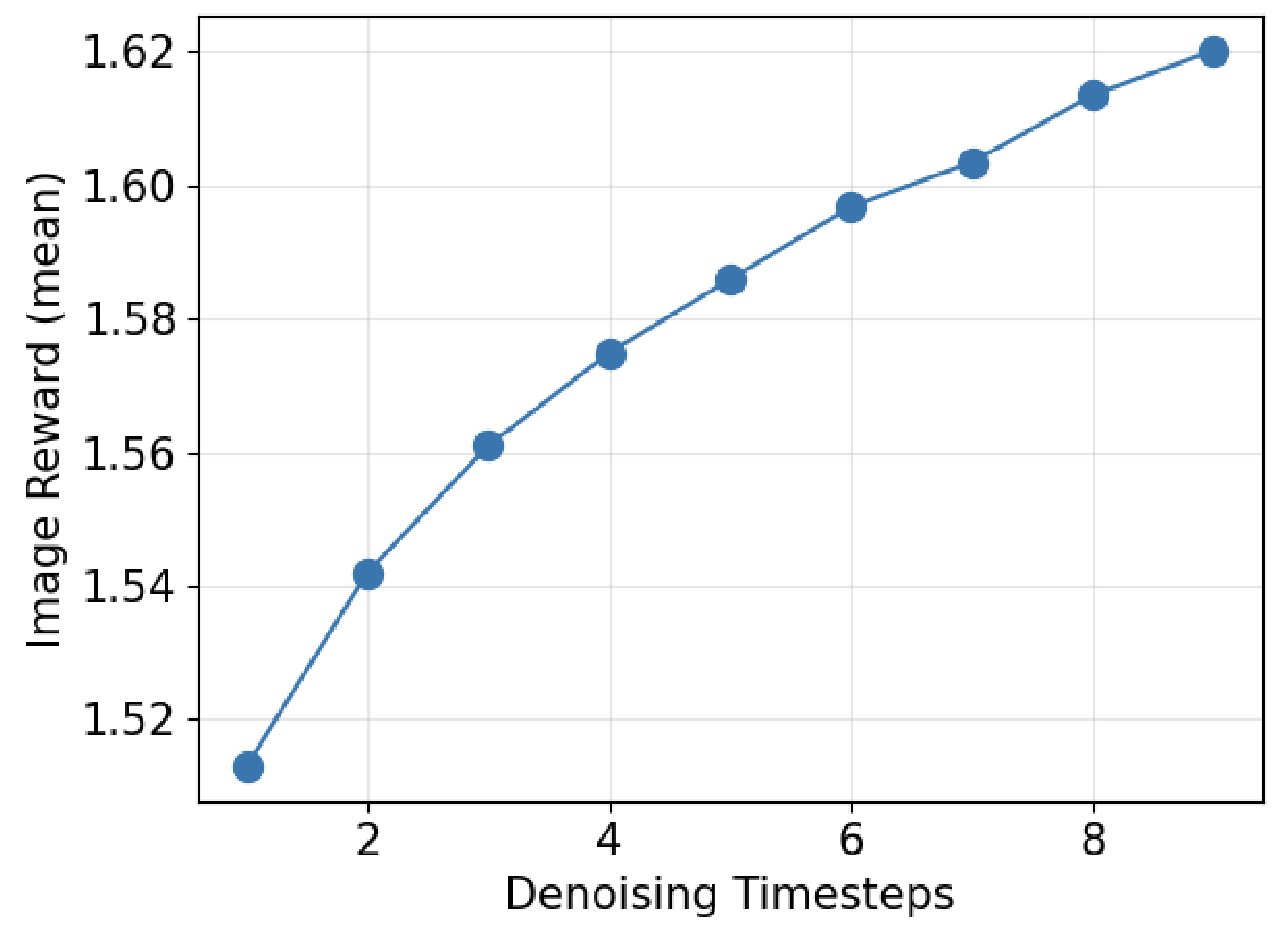} 
\caption{Verifier score variation for ImageReward used to derive $\Delta$.}
\label{fig: ablation_vt}
\end{minipage}
\end{figure}
\vspace{-5pt}
\section{Related Works}
\label{app: related work}
\vspace{-5pt}
\subsection*{Test-time compute as a new scaling axis}
\vspace{-5pt}
In recent years, scaling of AI models has mainly involved increasing model size and the amount of compute and data utilized during training. However, recent work shows that spending more compute at test-time can substantially improve output quality and reasoning capability, thus creating a new dimension for scaling compute to improve model performance. OpenAI's o1 series popularized this paradigm by spending more test-time budget on deliberate “thinking” before answering~\cite{openai2024reason}. DeepSeek-R1/R1-Zero similarly leverages extended reasoning trajectories at inference~\cite{guo2025deepseek}. Snell et al.\ formalize when scaling test-time compute can be more effective than scaling model size~\cite{snell2024scaling}. Complementing these analyses, \emph{s1} demonstrates a simple and effective test-time scaling recipe with strong empirical gains~\cite{muennighoff2025s1}. 
\vspace{-10pt}
\subsection*{Test-time scaling for image generation}
\vspace{-5pt}
While diffusion models and flow models have achieved impressive results in image and video generation, naive generation with these models often fails to satisfy complex user instructions involving object quantities, relative position, and size, among other conceptual attributes. Recently, test-time scaling methods have been developed for diffusion and flow models that significantly outperform naive generation. 
\vspace{-10pt}
\subsubsection*{Reward-based sampling and noise-search}
\vspace{-5pt}
Reward-based sampling methods repeatedly sample from the model’s learned distribution, guided by a reward such as text–image alignment or aesthetic quality, to select higher-quality outputs. The most basic example is Best-of-$N$ (BoN), which generates $N$ independent images with different random seeds and picks the one with the highest reward. Ma et al.\ (2025) formalize this paradigm by introducing both \textbf{BoN} and Search-over-Paths (\textbf{SoP}), the latter refining trajectories by sampling particles along the denoising path~\cite{ma2025inference}. Building on this idea, “noise trajectory search” methods incorporate reward-guided selection directly into the denoising process: \textbf{SVDD}~\cite{svdd} selects, at every denoising step, the particle with the highest expected reward.
Extending reward-based search to flow models, Kim et al.\ (2025)~\cite{kim2025inference} propose three complementary techniques: test-time SDE conversion, which introduces stochasticity into otherwise deterministic flows; interpolant conversion, which alters the trajectory interpolant to expand the search space; and Rollover Budget Forcing (\textbf{RBF}), which dynamically reallocates compute across timesteps by advancing any particle that exceeds the previous best reward and rolling over unused function evaluations to future steps, thereby ensuring efficient utilization of the total budget.


\vspace{-5pt}
\begin{table}[ht]
\centering
\renewcommand{\arraystretch}{1.3} 
\setlength{\tabcolsep}{8pt}       
\begin{tabular}{|c|c|c|c|c|c|}
\hline
\textbf{Method} & \textbf{Generator} & \textbf{Verifier} & \textbf{NFEs} & \textbf{time ($\times$)} & \textbf{GenEval (\%)} \\
\hline
Regular-T2I & SDv2.1 & - & 1 & 1 & 47.93 \\
\hline
Regular-T2I & FLUX-Schnell & - & 10 & 4.74 & 62.33 \\
\hline
\hline
Steering & SDv2.1 & ImageReward & 4 & 3.57 & 57.93 \\
Steering & SDv2.1 & ImageReward & 20 & 15.62 & 64.91 \\
Steering & SDv2.1 & ImageReward & 40 & 31.91 & 66.18 \\
\hline
\hline
BoN & FLUX-Schnell & ImageReward & 40 & 9.70 & 70.62 \\
BoN & FLUX-Schnell & ImageReward & 80 & 17.16 & 70.36 \\
BoN & FLUX-Schnell & ImageReward & 160 & 26.39 & 72.45 \\
\hline
BoN & FLUX-Schnell & VQAScore & 40 & 10.29 & 70.07 \\
BoN & FLUX-Schnell & VQAScore & 80 & 18.72 & 70.54 \\
BoN & FLUX-Schnell & VQAScore & 160 & 27.07 & 71.91 \\
\hline
\hline
SoP & FLUX-Schnell & ImageReward & 250 & 21.00 & 60.68 \\
\hline
SoP & FLUX-Schnell & VQAScore & 250 & 22.25 & 61.19 \\
\hline
\hline
SVDD & FLUX-Schnell & ImageReward & 250 & 43.69 & 68.76 \\
\hline
SVDD & FLUX-Schnell & VQAScore & 250 & 56.74 & 72.89 \\
\hline
\hline
FlowRBF & FLUX-Schnell & ImageReward & 40 & 12.80 & 70.10 \\
FlowRBF & FLUX-Schnell & ImageReward & 80 & 18.11 & 72.45 \\
FlowRBF & FLUX-Schnell & ImageReward & 160 & 34.11 & 74.20 \\
\hline
FlowRBF & FLUX-Schnell & VQAScore & 40 & 13.83 & 70.32 \\
FlowRBF & FLUX-Schnell & VQAScore & 80 & 27.03 & 71.77 \\
FlowRBF & FLUX-Schnell & VQAScore & 160 & 48.98 & 73.45 \\
\hline
\hline
VT (Ours) & FLUX-Schnell & ImageReward & 40 & 10.31 & 71.61 \\
VT (Ours) & FLUX-Schnell & ImageReward & 80 & 18.68 & 76.77 \\
VT (Ours) & FLUX-Schnell & ImageReward & 160 & 39.36 & \textbf{78.62} \\
\hline
VT (Ours) & FLUX-Schnell & VQAScore & 40 & 12.80 & 73.83 \\
VT (Ours) & FLUX-Schnell & VQAScore & 80 & 25.90 & 76.12 \\
VT (Ours) & FLUX-Schnell & VQAScore & 160 & 51.97 & 77.20 \\
\hline
\end{tabular}
\vspace{5pt}
\caption{Comparison of different approaches across models, verifiers, NFEs, along with their efficiency (wall-clock time) \& performance (GenEval) scores. Methods are Steering~\cite{singhal2025general}, Best-of-$N$~\cite{ma2025inference} (BoN), Search-over-Paths~\cite{ma2025inference} (SoP), Soft Value-based Decoding in Diffusion~\cite{li2024derivative} (SVDD), FlowRBF~\cite{kim2025inference}, and Verifier-Threshold (VT). Verifiers are ImageReward~\cite{xu2023imagereward} and VQAScore~\cite{lin2024evaluating}.}
\label{tab: full results}
\end{table}

\section{Comprehensive Results}
\label{app: results}
\vspace{-10pt}
In this section, we expand the plots in Fig.~\ref{fig: intro} and Fig.~\ref{fig: results image reward} into Tab.~\ref{tab: full results}. We also show images for various example prompts, including cases where our approach succeeds while other baselines fail, as well as cases where all approaches fail. In all example triplets, Fig.~(a) corresponds to generation without test-time scaling, Fig.~(b) corresponds to FlowRBF, and Fig.~(c) corresponds to our \textit{Verifier-Threshold} algorithm.

\begin{figure}[th]
    \centering
    \begin{subfigure}{0.30\linewidth}
        \centering
        \includegraphics[width=\linewidth]{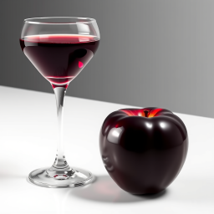}
        \caption{}
    \end{subfigure}
    \hfill
    \begin{subfigure}{0.30\linewidth}
        \centering
        \includegraphics[width=\linewidth]{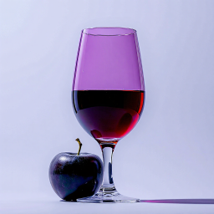}
        \caption{}
    \end{subfigure}
    \hfill
    \begin{subfigure}{0.30\linewidth}
        \centering
        \includegraphics[width=\linewidth]{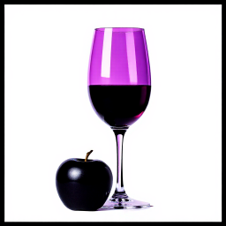}
        \caption{}
    \end{subfigure}

    \caption{``a photo of a purple wine glass and a black apple''. For all examples in this section, Fig. (a) is the image generated without test-time scaling, Fig. (b) is generated using FlowRBF and Fig. (c) is generated by our Verifier-Threshold algorithm.}
    \label{fig: example1}
\end{figure}

\begin{figure}[th]
    \centering
    \begin{subfigure}{0.30\linewidth}
        \centering
        \includegraphics[width=\linewidth]{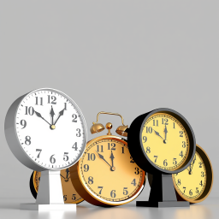}
        \caption{}
    \end{subfigure}
    \hfill
    \begin{subfigure}{0.30\linewidth}
        \centering
        \includegraphics[width=\linewidth]{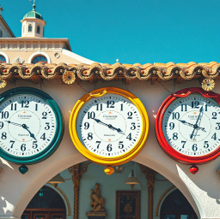}
        \caption{}
    \end{subfigure}
    \hfill
    \begin{subfigure}{0.30\linewidth}
        \centering
        \includegraphics[width=\linewidth]{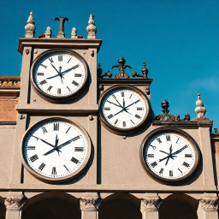}
        \caption{}
    \end{subfigure}

    \caption{``a photo of four clocks''}
    \label{fig: example2}
\end{figure}

\begin{figure}[th]
    \centering
    \begin{subfigure}{0.30\linewidth}
        \centering
        \includegraphics[width=\linewidth]{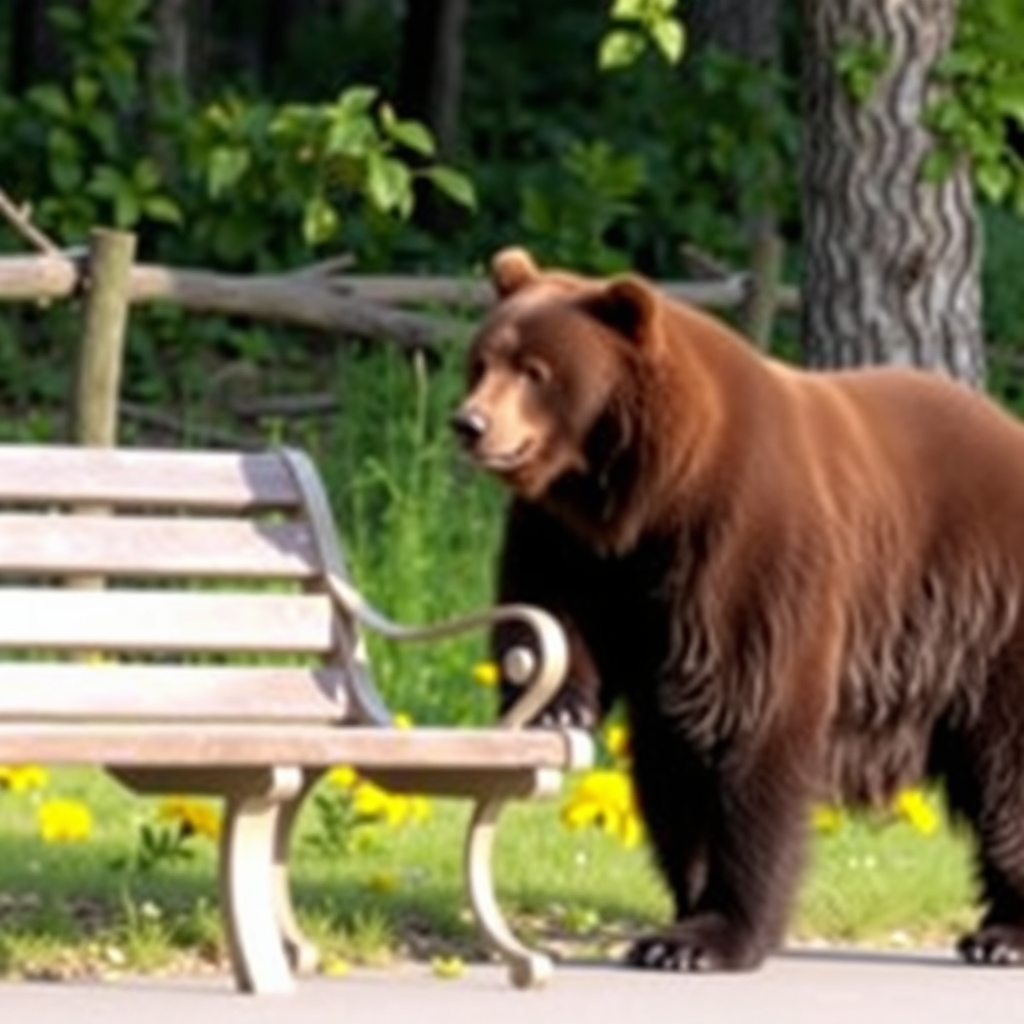}
        \caption{}
    \end{subfigure}
    \hfill
    \begin{subfigure}{0.30\linewidth}
        \centering
        \includegraphics[width=\linewidth]{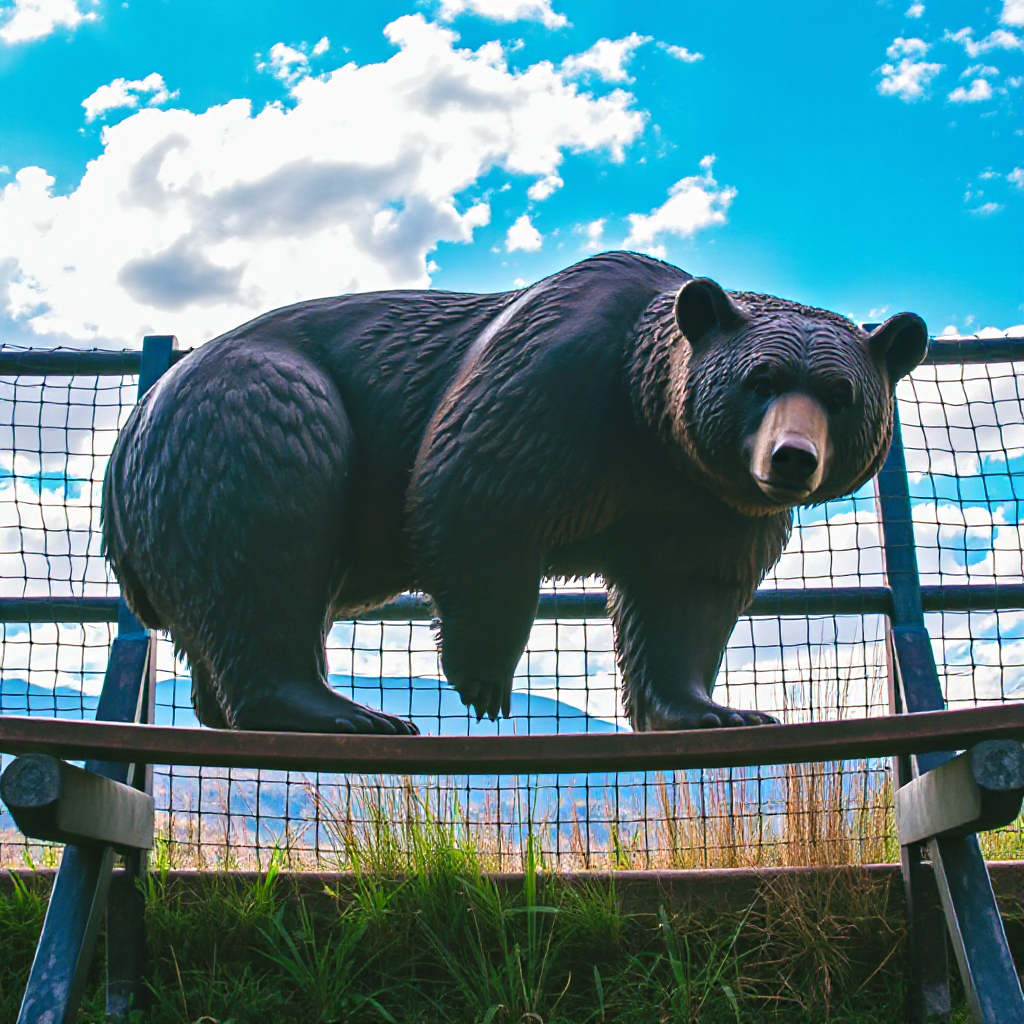}
        \caption{}
    \end{subfigure}
    \hfill
    \begin{subfigure}{0.30\linewidth}
        \centering
        \includegraphics[width=\linewidth]{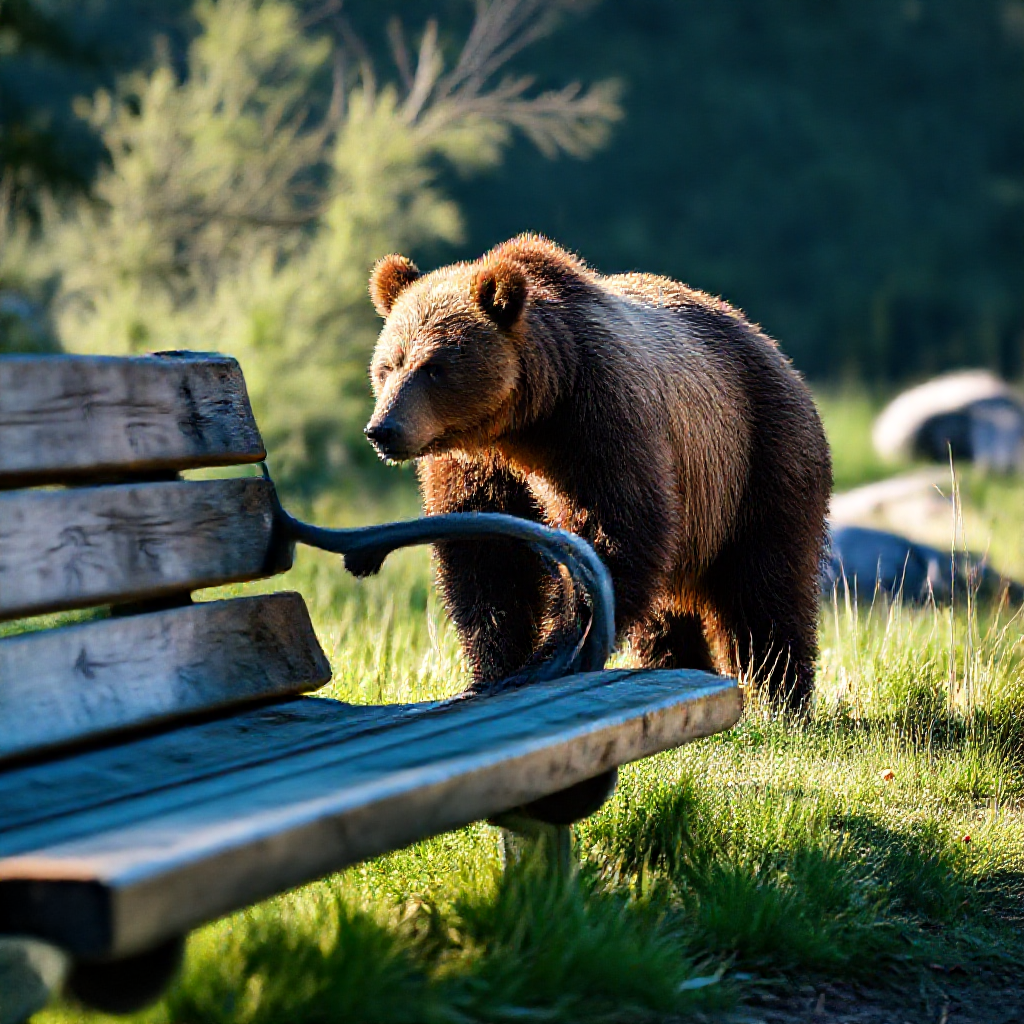}
        \caption{}
    \end{subfigure}

    \caption{``a photo of a bench left of a bear''}
    \label{fig: example3}
\end{figure}

\begin{figure}[th]
    \centering
    \begin{subfigure}{0.30\linewidth}
        \centering
        \includegraphics[width=\linewidth]{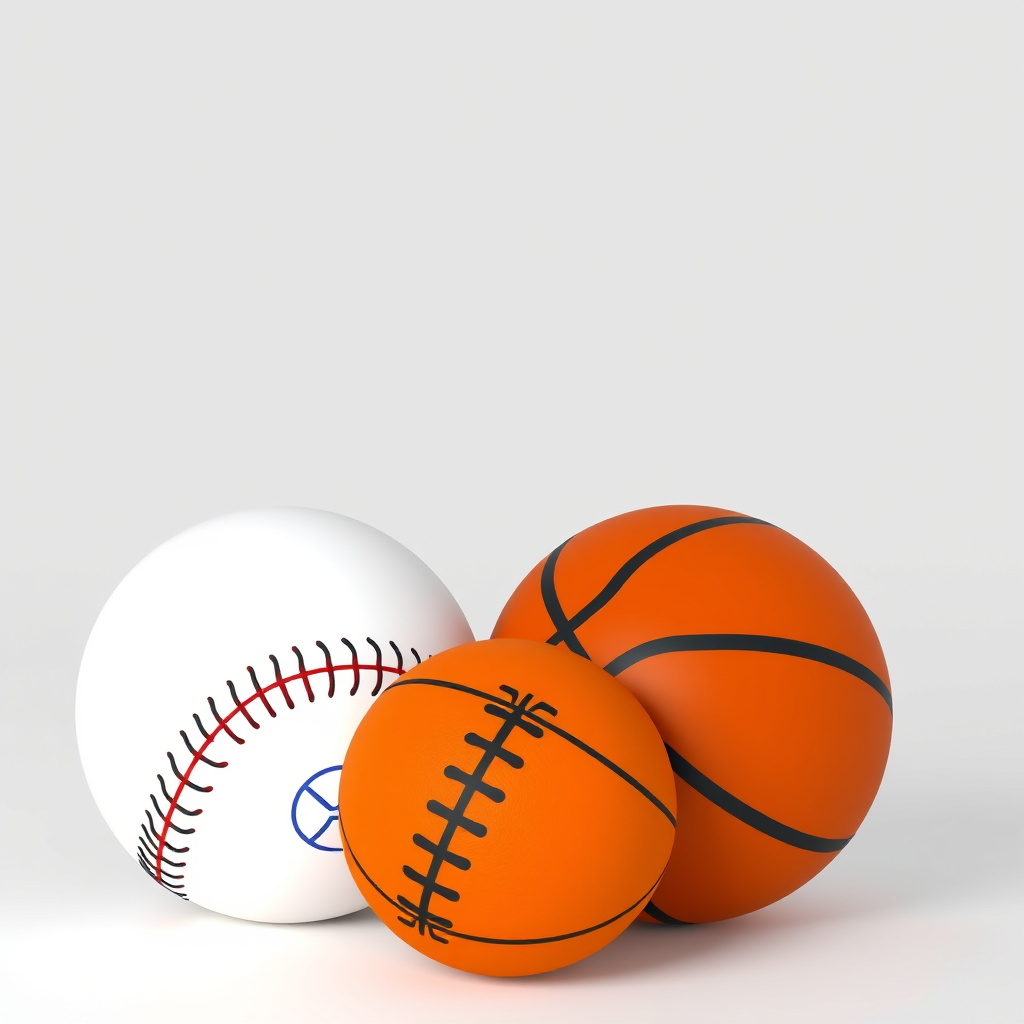}
        \caption{}
    \end{subfigure}
    \hfill
    \begin{subfigure}{0.30\linewidth}
        \centering
        \includegraphics[width=\linewidth]{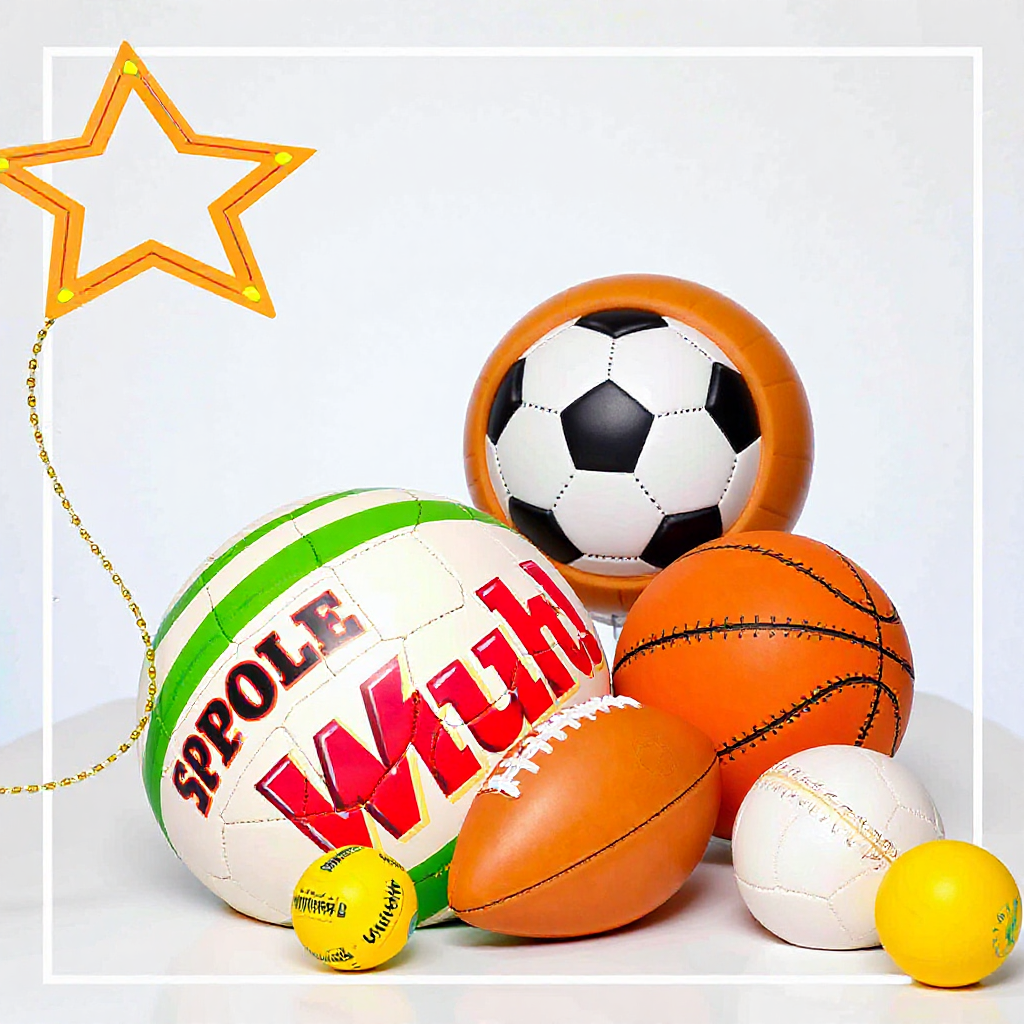}
        \caption{}
    \end{subfigure}
    \hfill
    \begin{subfigure}{0.30\linewidth}
        \centering
        \includegraphics[width=\linewidth]{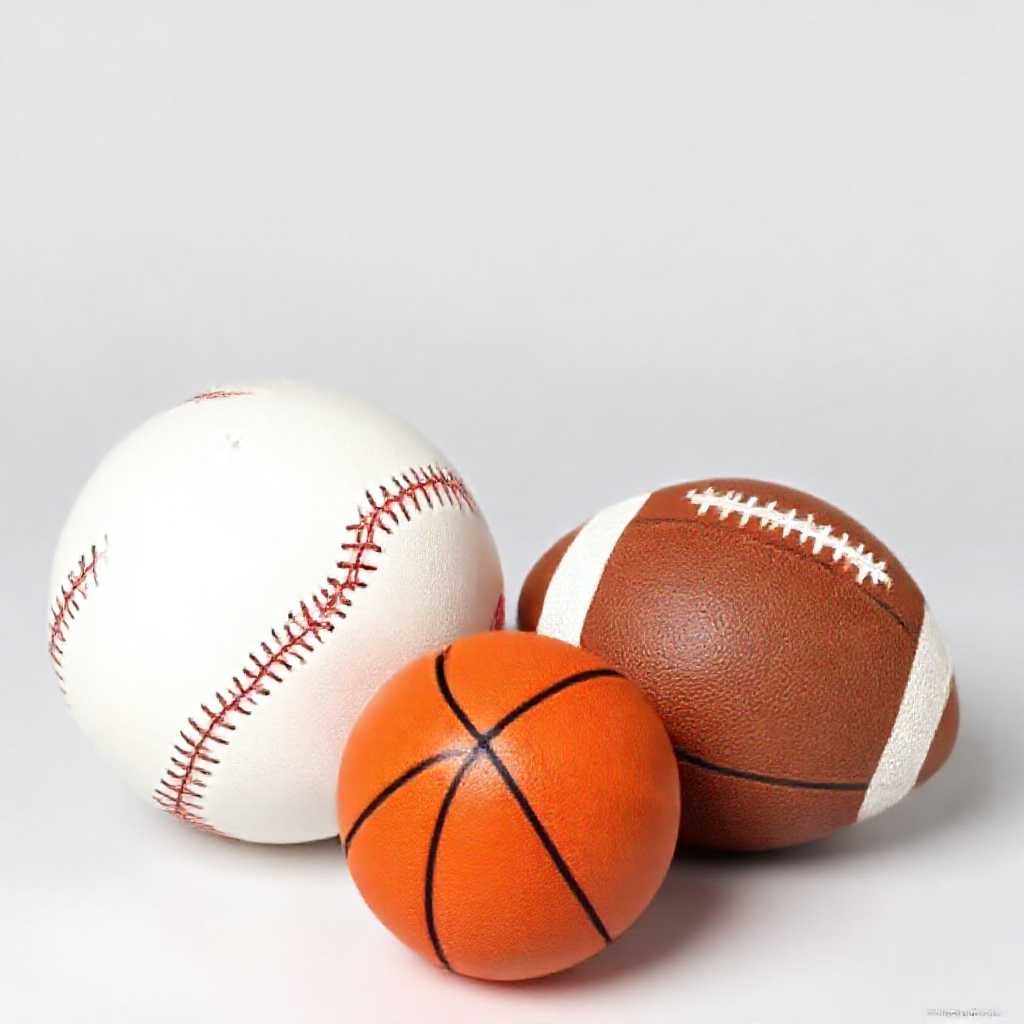}
        \caption{}
    \end{subfigure}

    \caption{``a photo of three sports balls''}
    \label{fig: example4}
\end{figure}

\begin{figure}[th]
    \centering
    \begin{subfigure}{0.30\linewidth}
        \centering
        \includegraphics[width=\linewidth]{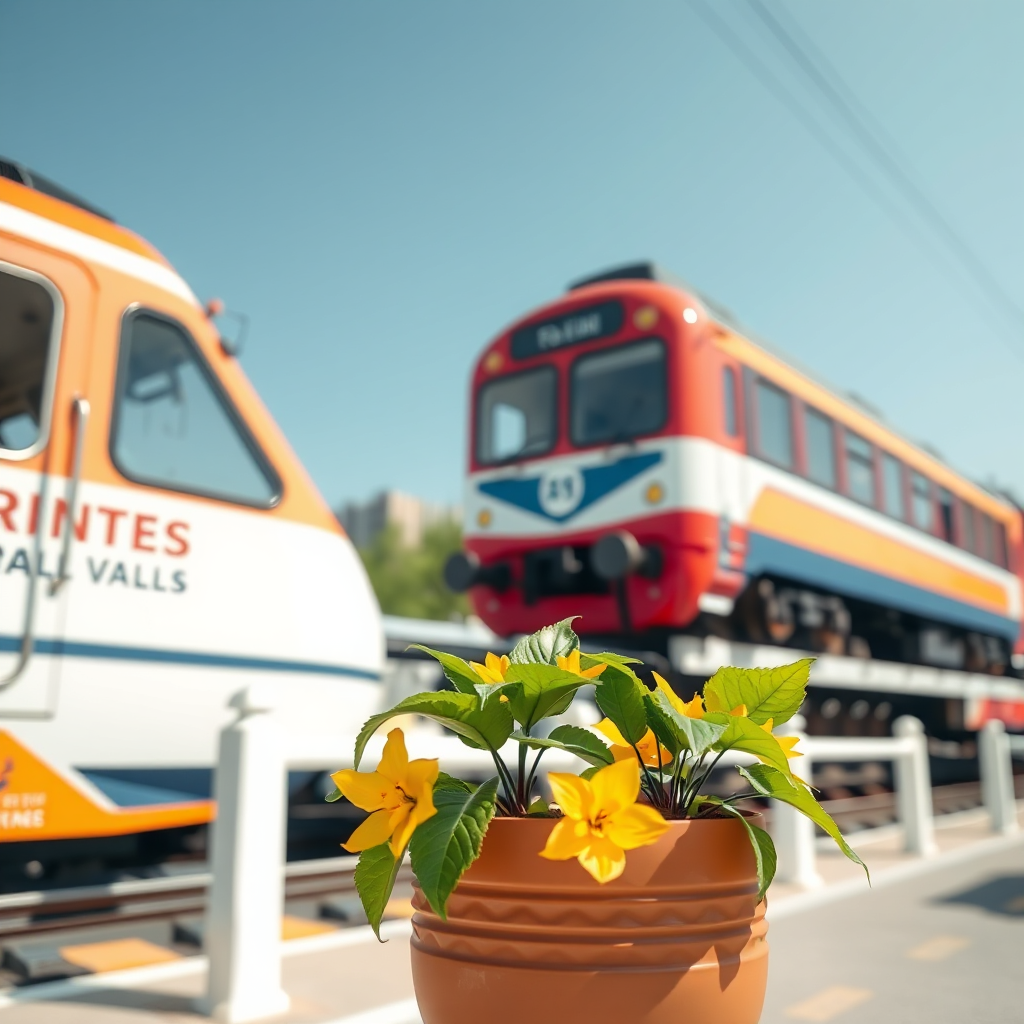}
        \caption{}
    \end{subfigure}
    \hfill
    \begin{subfigure}{0.30\linewidth}
        \centering
        \includegraphics[width=\linewidth]{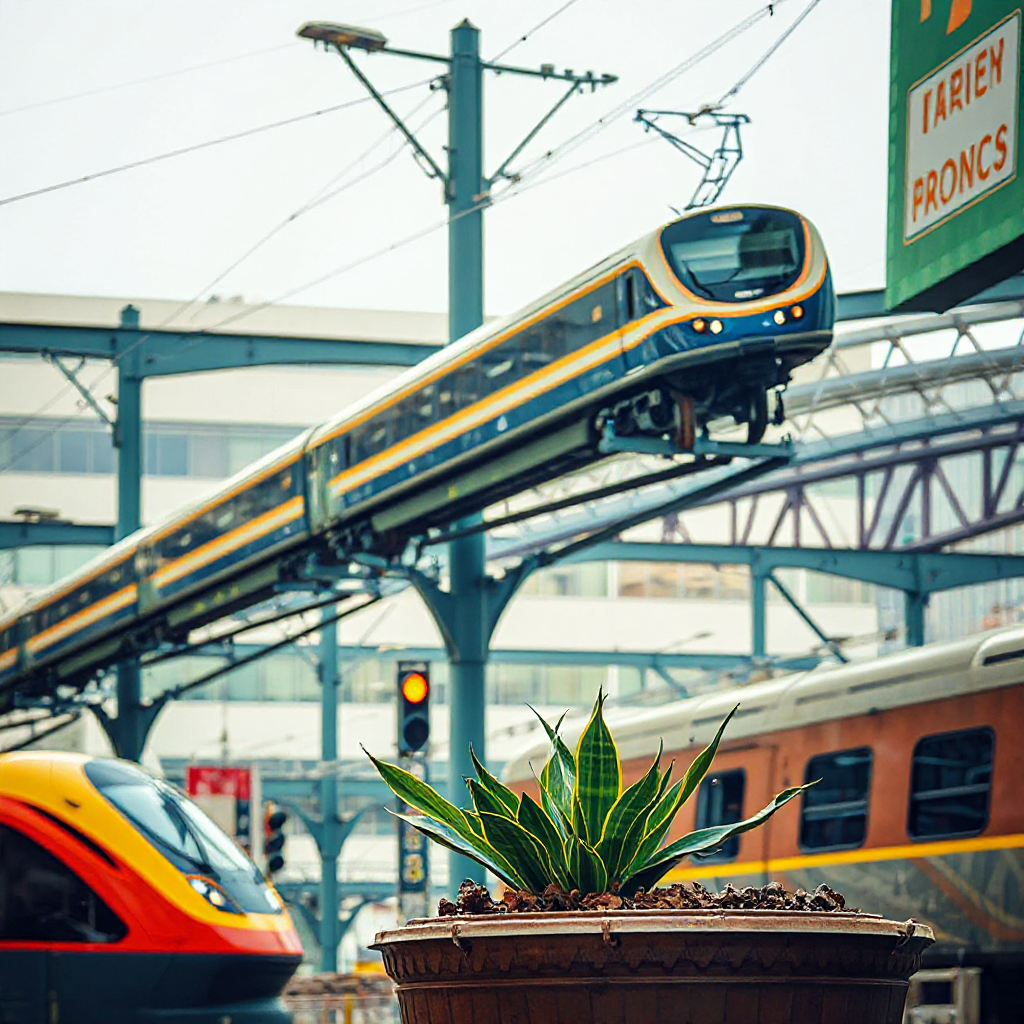}
        \caption{}
    \end{subfigure}
    \hfill
    \begin{subfigure}{0.30\linewidth}
        \centering
        \includegraphics[width=\linewidth]{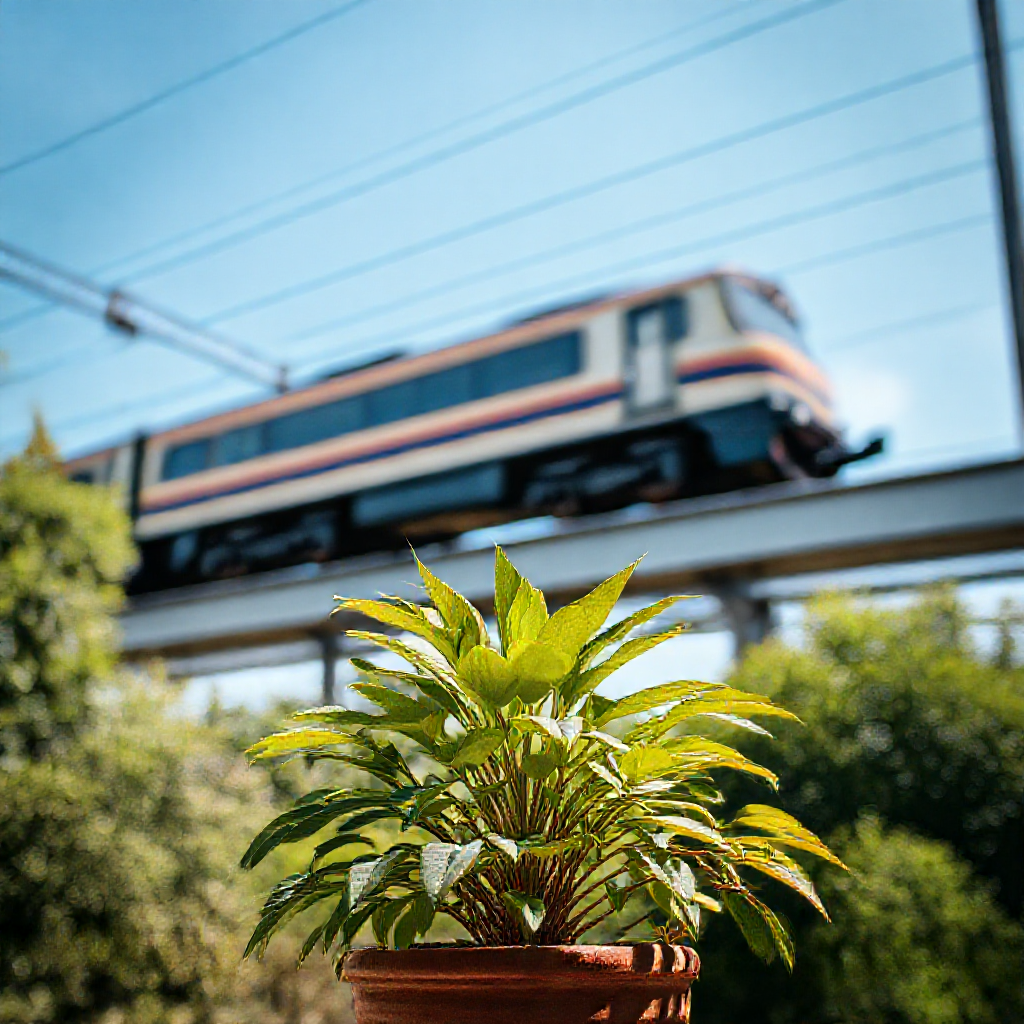}
        \caption{}
    \end{subfigure}

    \caption{``a photo of a train above a potted plant''}
    \label{fig: example5}
\end{figure}

\begin{figure}[th]
    \centering
    \begin{subfigure}{0.30\linewidth}
        \centering
        \includegraphics[width=\linewidth]{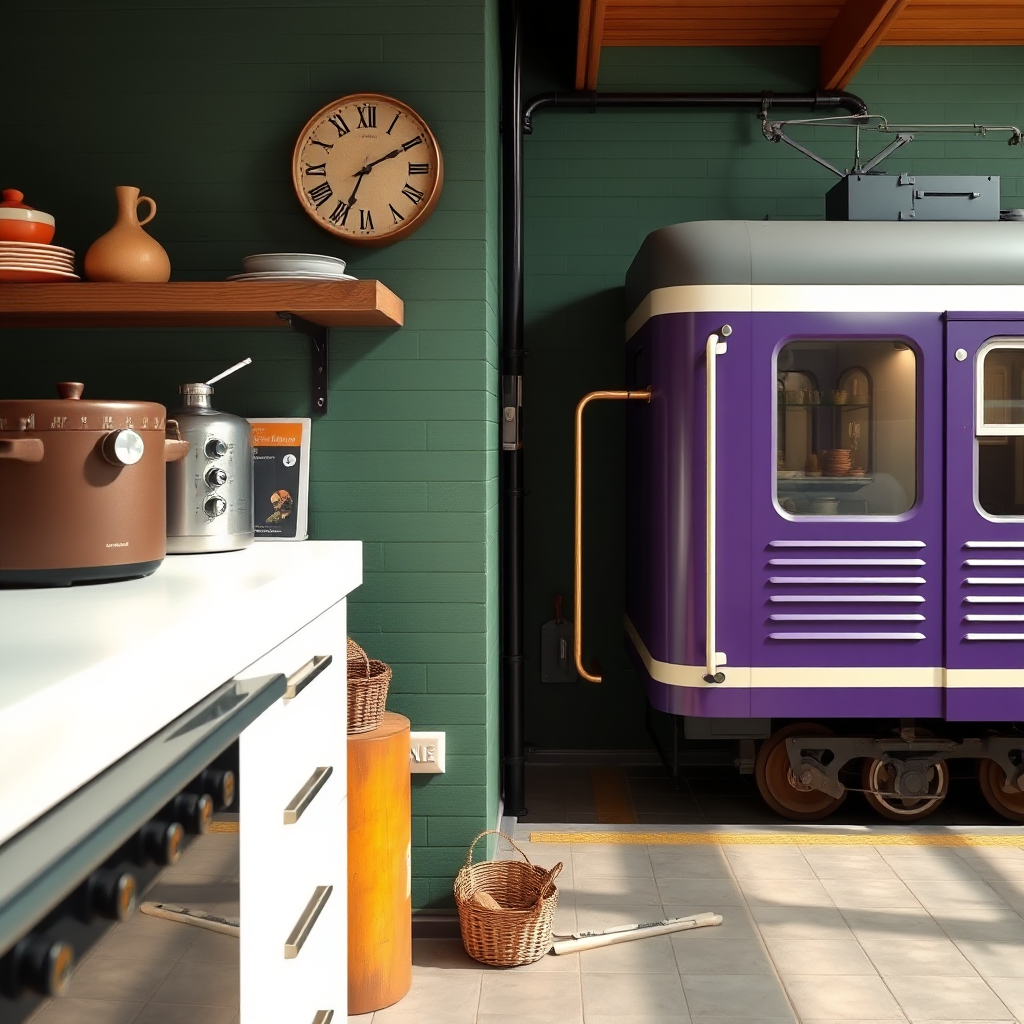}
        \caption{}
    \end{subfigure}
    \hfill
    \begin{subfigure}{0.30\linewidth}
        \centering
        \includegraphics[width=\linewidth]{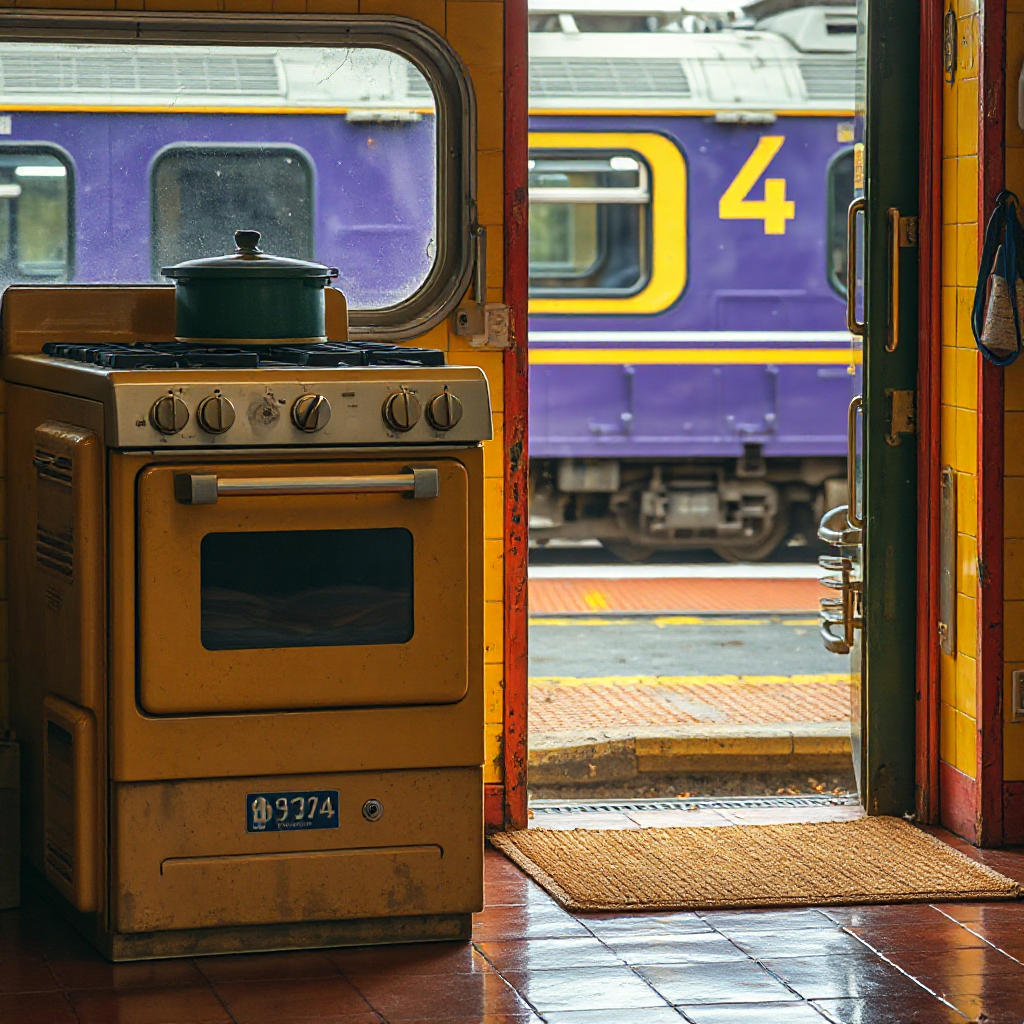}
        \caption{}
    \end{subfigure}
    \hfill
    \begin{subfigure}{0.30\linewidth}
        \centering
        \includegraphics[width=\linewidth]{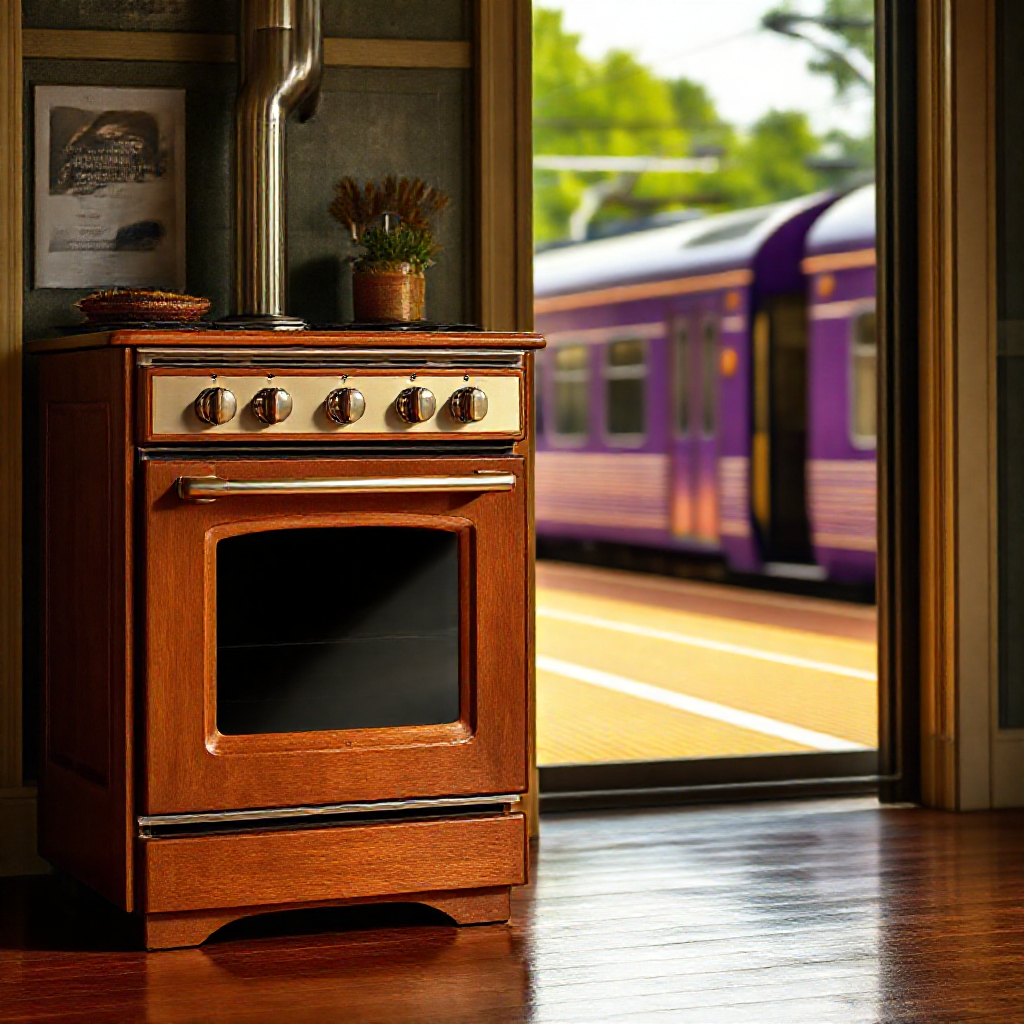}
        \caption{}
    \end{subfigure}

    \caption{``a photo of a brown oven and a purple train''}
    \label{fig: example6}
\end{figure}



\begin{figure}[th]
    \centering
    \begin{subfigure}{0.30\linewidth}
        \centering
        \includegraphics[width=\linewidth]{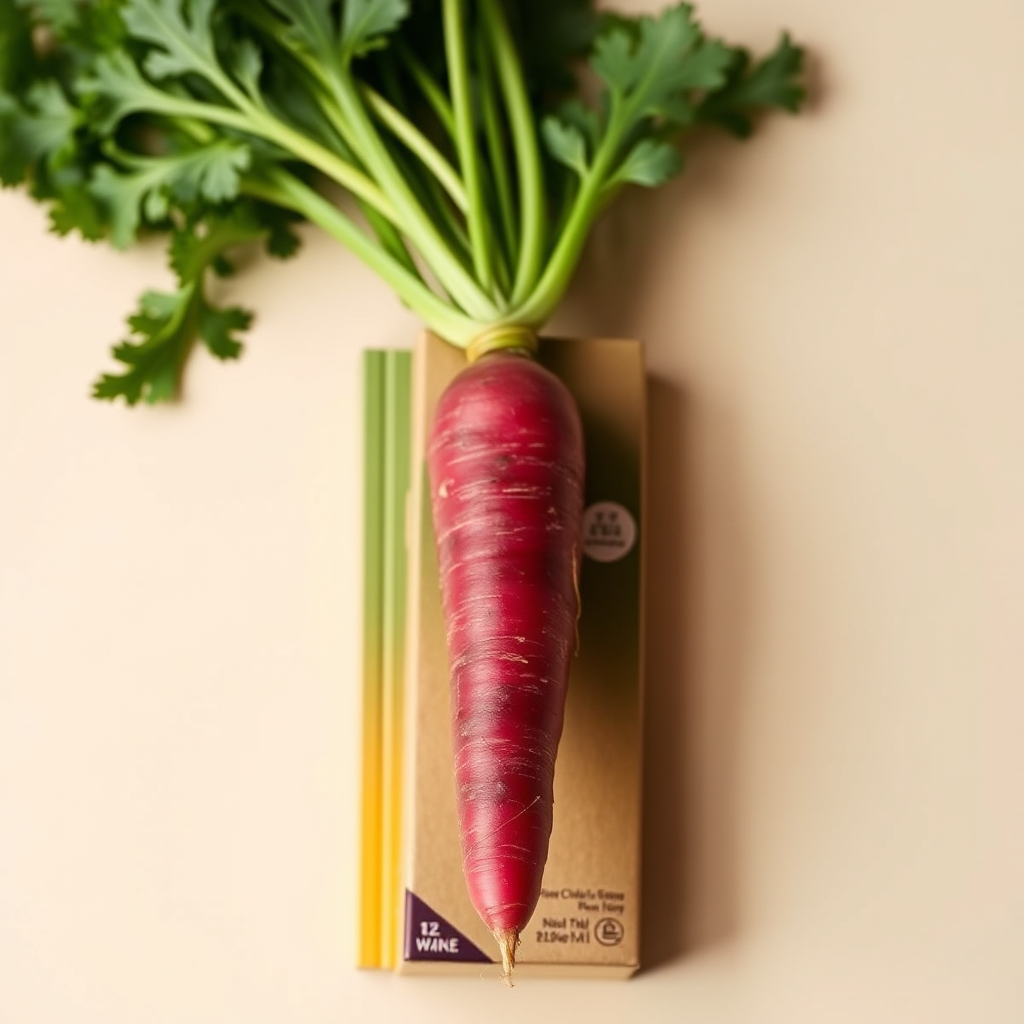}
        \caption{}
    \end{subfigure}
    \hfill
    \begin{subfigure}{0.30\linewidth}
        \centering
        \includegraphics[width=\linewidth]{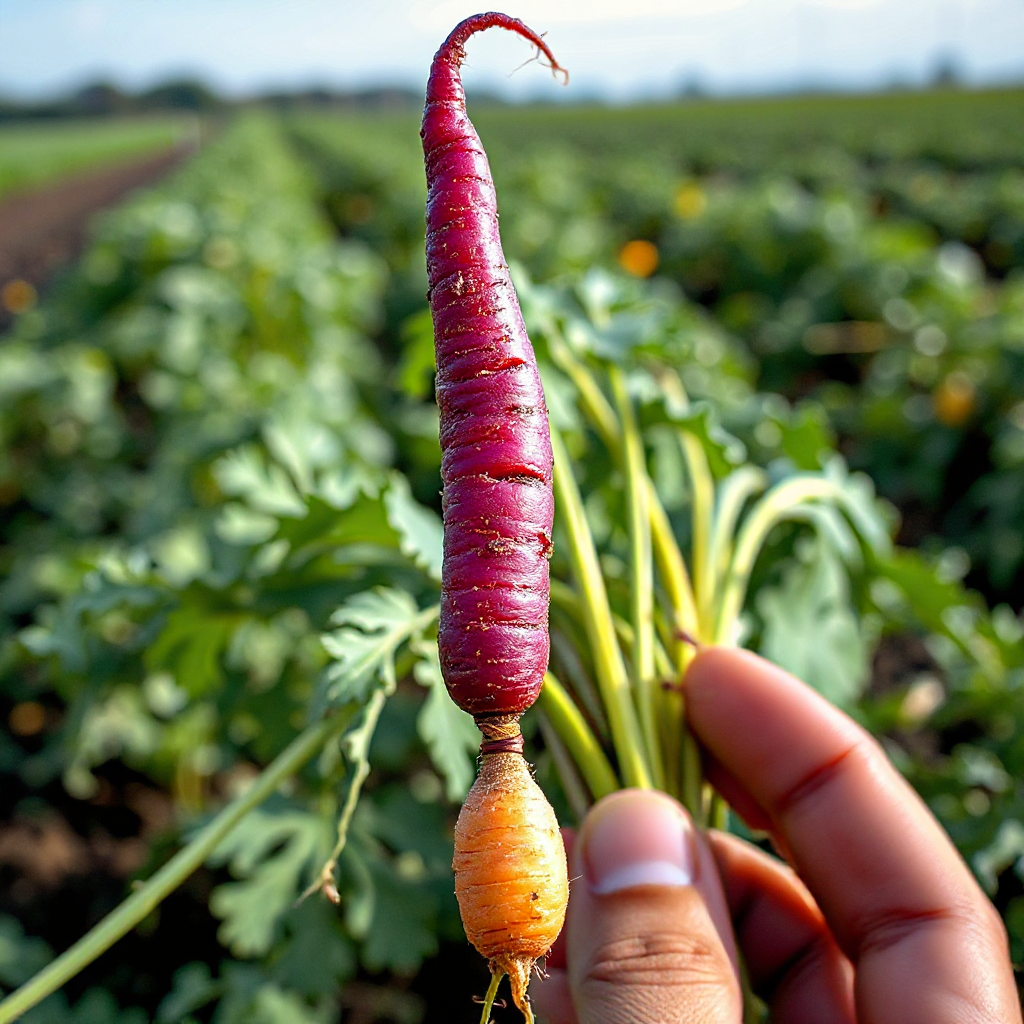}
        \caption{}
    \end{subfigure}
    \hfill
    \begin{subfigure}{0.30\linewidth}
        \centering
        \includegraphics[width=\linewidth]{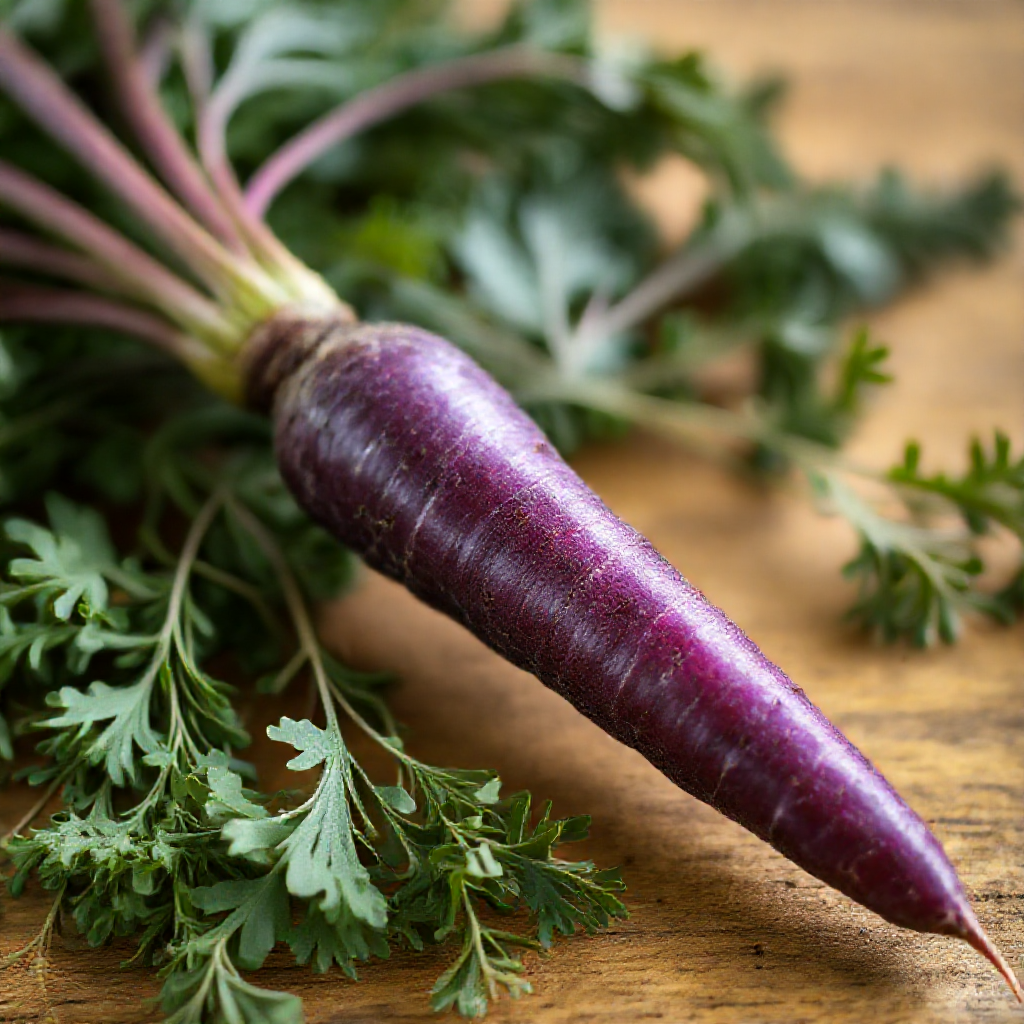}
        \caption{}
    \end{subfigure}

    \caption{``a photo of a purple carrot''}
    \label{fig: example8}
\end{figure}

\begin{figure}[th]
    \centering
    \begin{subfigure}{0.30\linewidth}
        \centering
        \includegraphics[width=\linewidth]{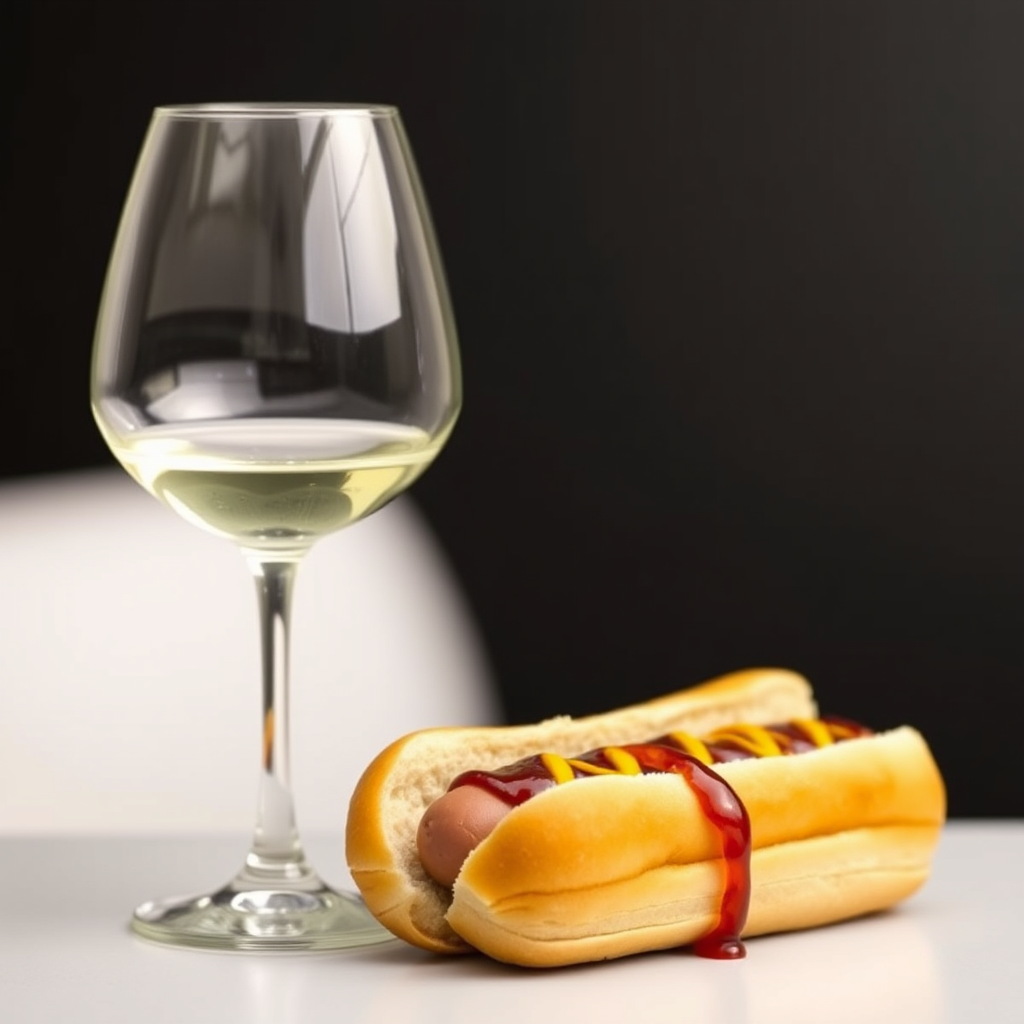}
        \caption{}
    \end{subfigure}
    \hfill
    \begin{subfigure}{0.30\linewidth}
        \centering
        \includegraphics[width=\linewidth]{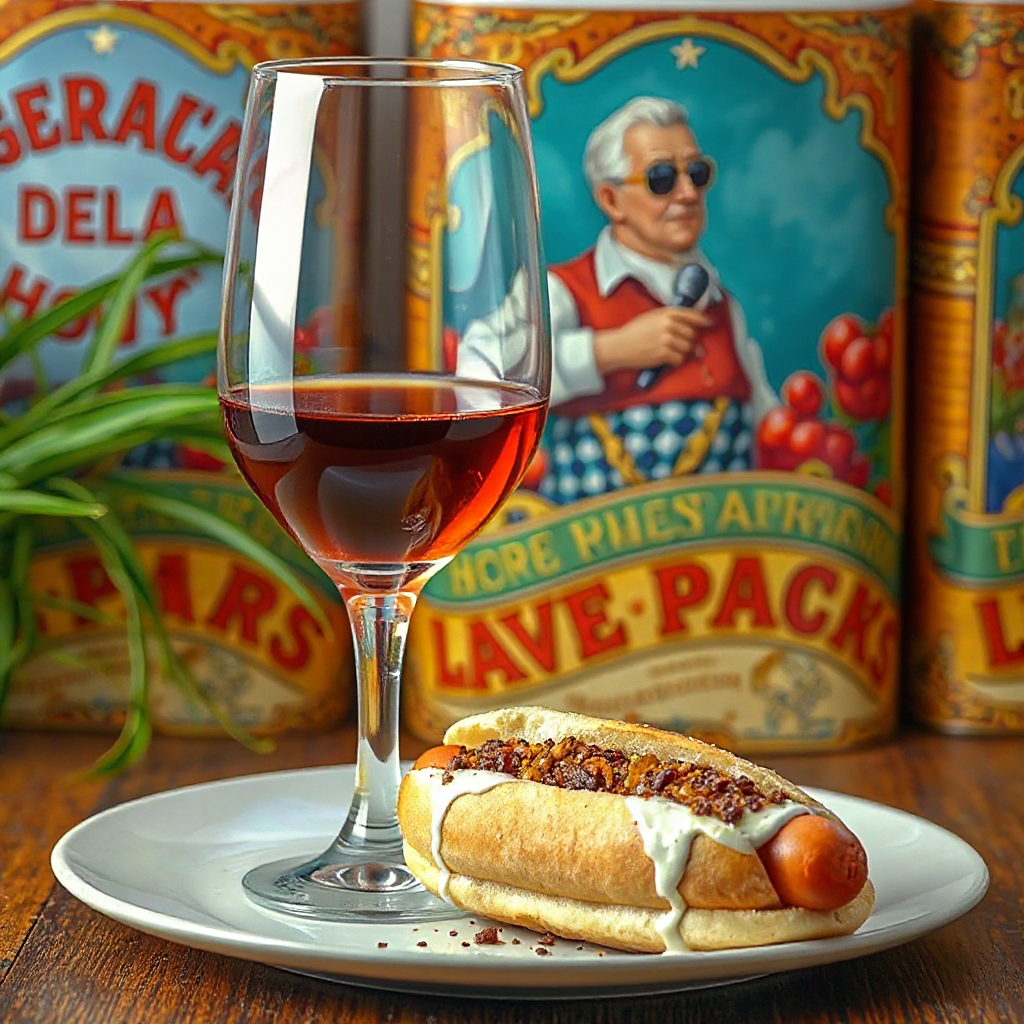}
        \caption{}
    \end{subfigure}
    \hfill
    \begin{subfigure}{0.30\linewidth}
        \centering
        \includegraphics[width=\linewidth]{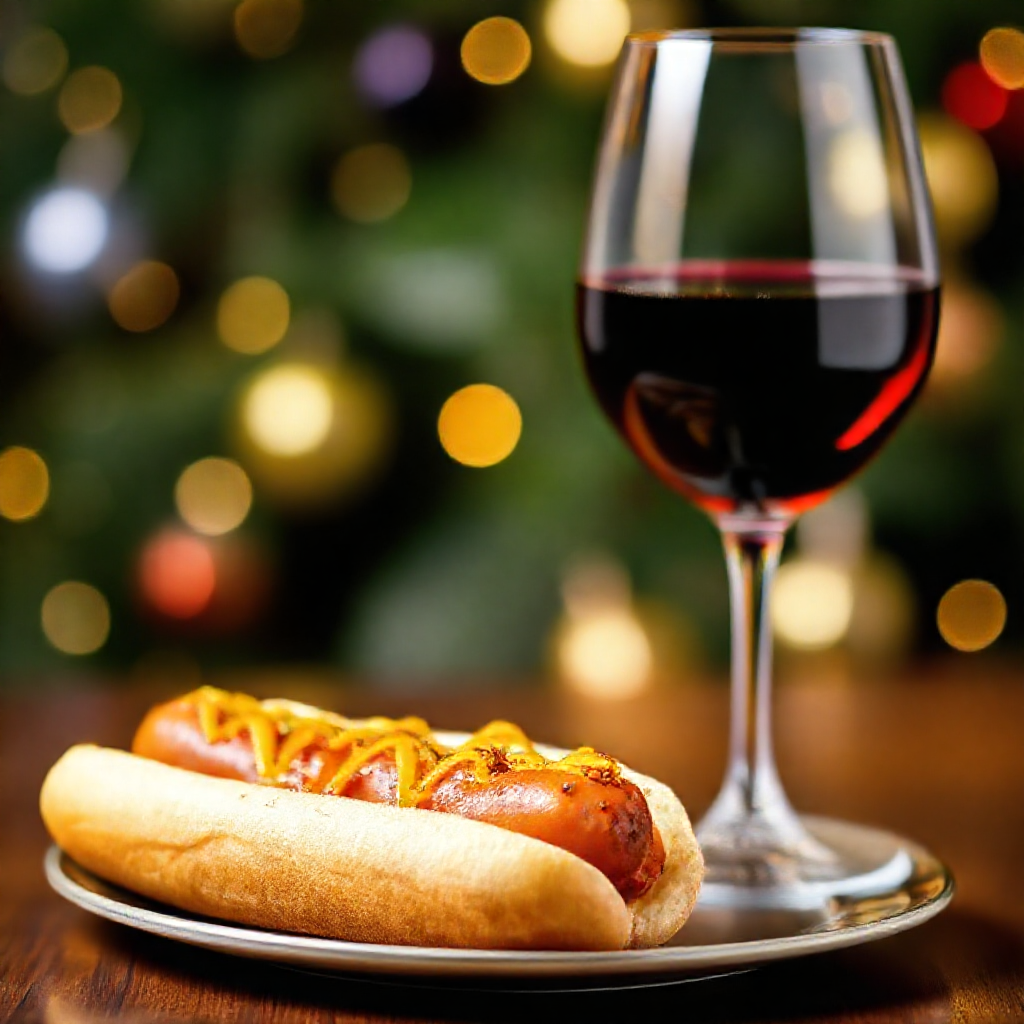}
        \caption{}
    \end{subfigure}

    \caption{``a photo of a wine glass right of a hot dog''}
    \label{fig: example9}
\end{figure}

\begin{figure}[th]
    \centering
    \begin{subfigure}{0.30\linewidth}
        \centering
        \includegraphics[width=\linewidth]{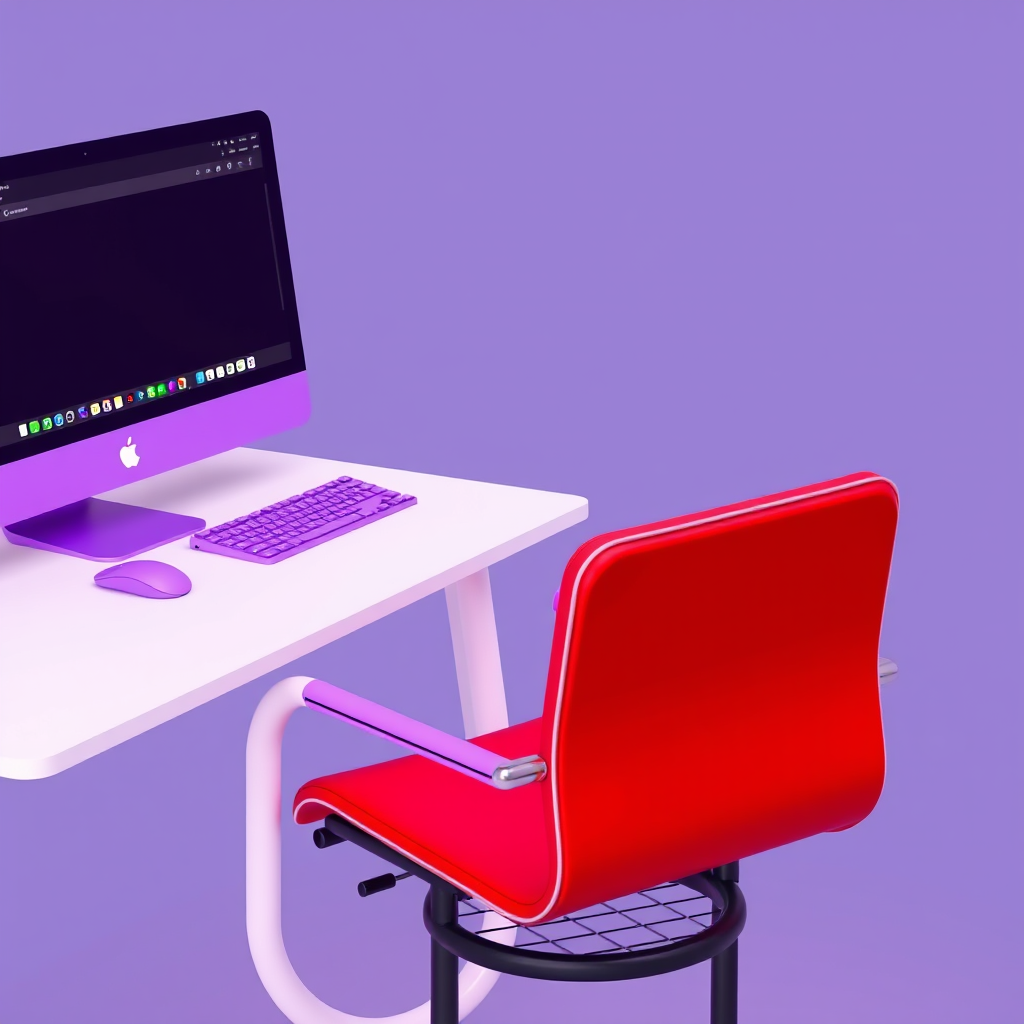}
        \caption{}
    \end{subfigure}
    \hfill
    \begin{subfigure}{0.30\linewidth}
        \centering
        \includegraphics[width=\linewidth]{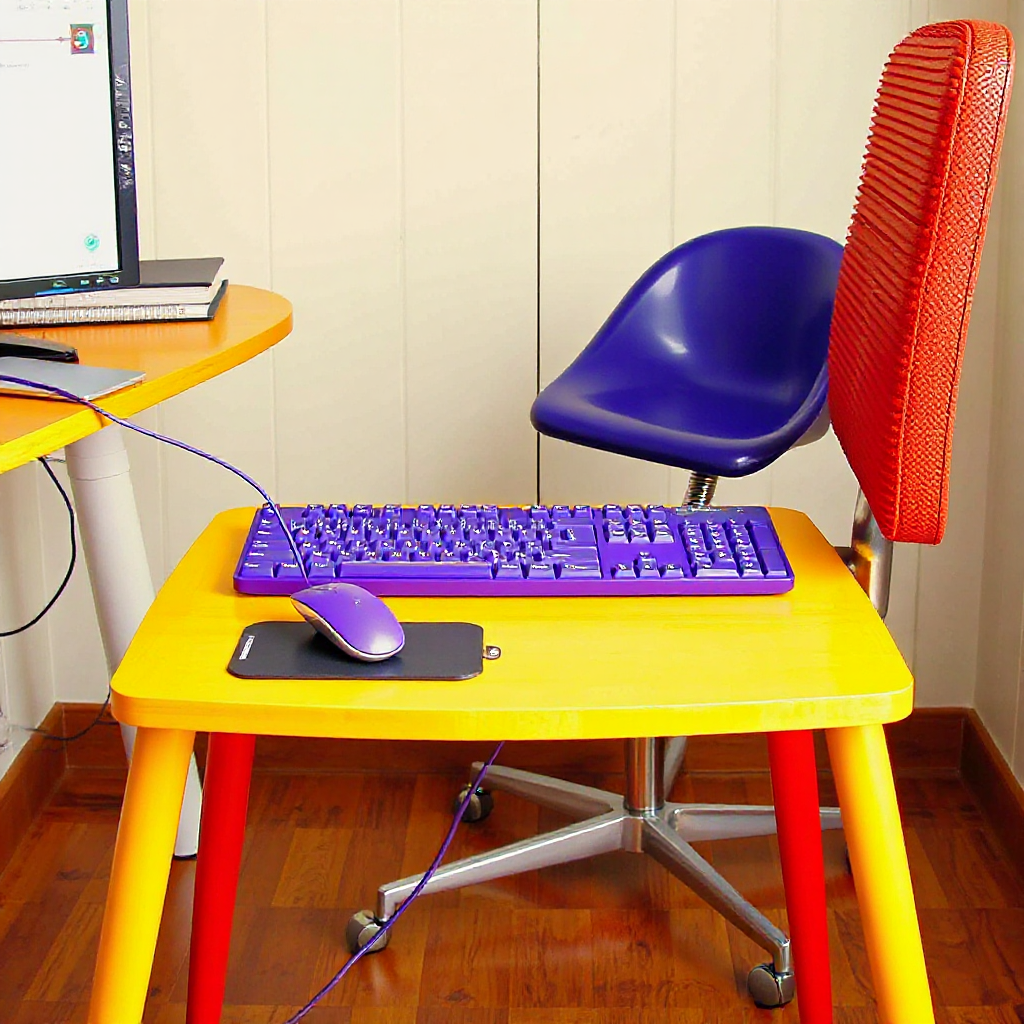}
        \caption{}
    \end{subfigure}
    \hfill
    \begin{subfigure}{0.30\linewidth}
        \centering
        \includegraphics[width=\linewidth]{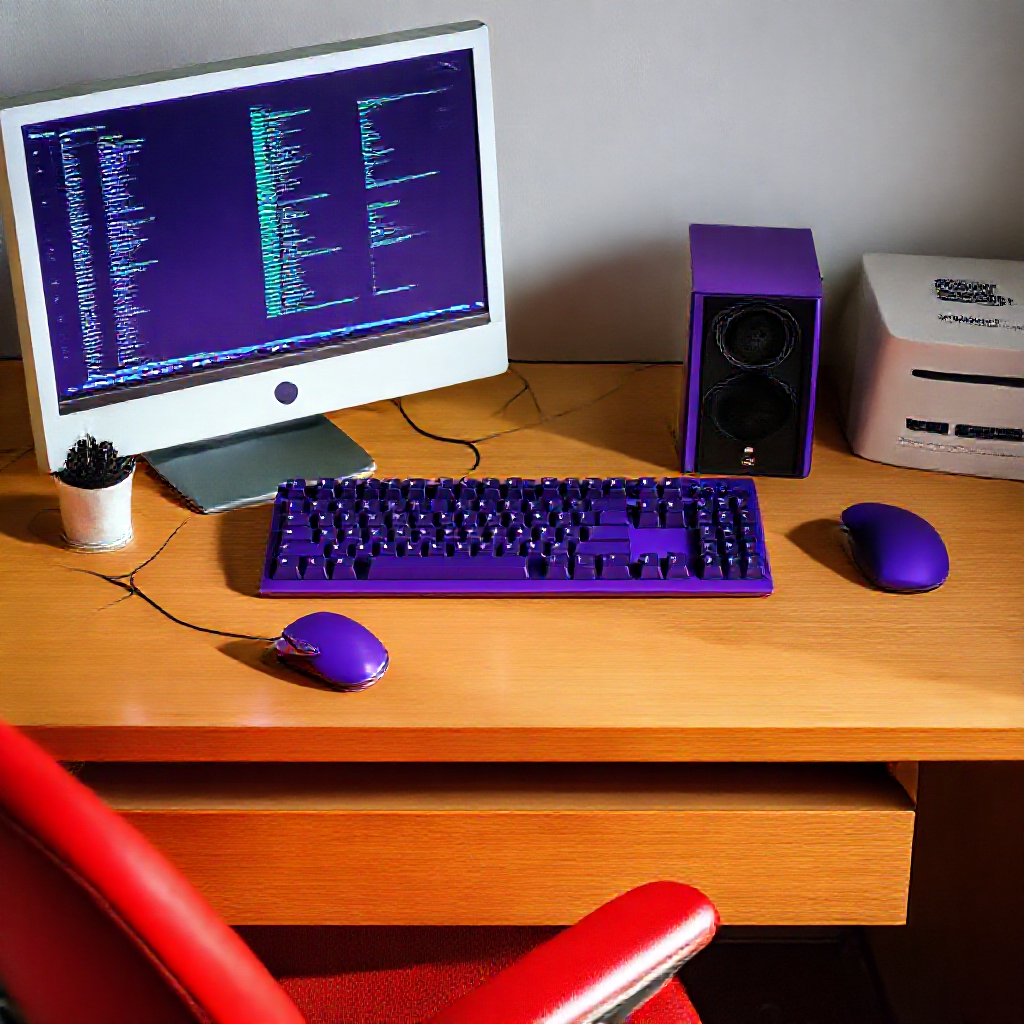}
        \caption{}
    \end{subfigure}

    \caption{``a photo of a purple computer keyboard and a red chair''}
    \label{fig: example10}
\end{figure}

\begin{figure}[th]
    \centering
    \begin{subfigure}{0.30\linewidth}
        \centering
        \includegraphics[width=\linewidth]{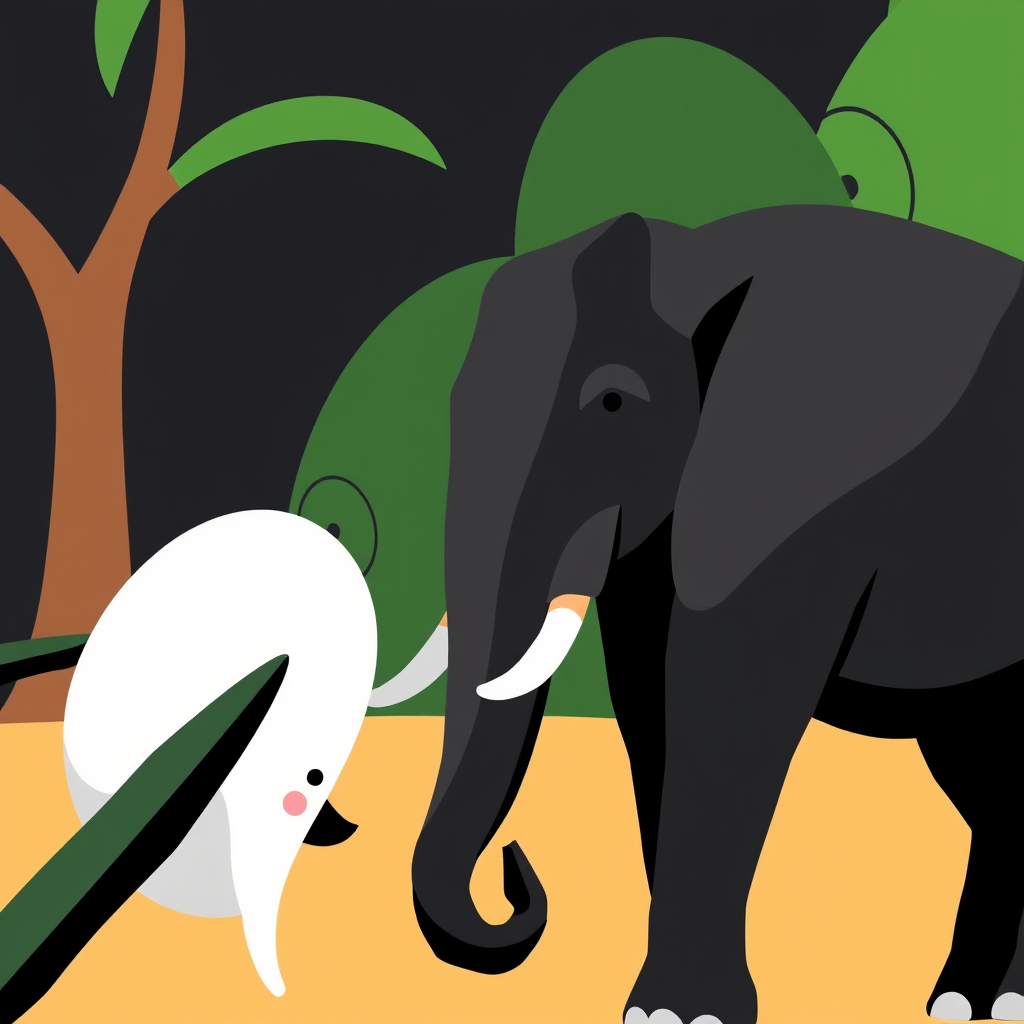}
        \caption{}
    \end{subfigure}
    \hfill
    \begin{subfigure}{0.30\linewidth}
        \centering
        \includegraphics[width=\linewidth]{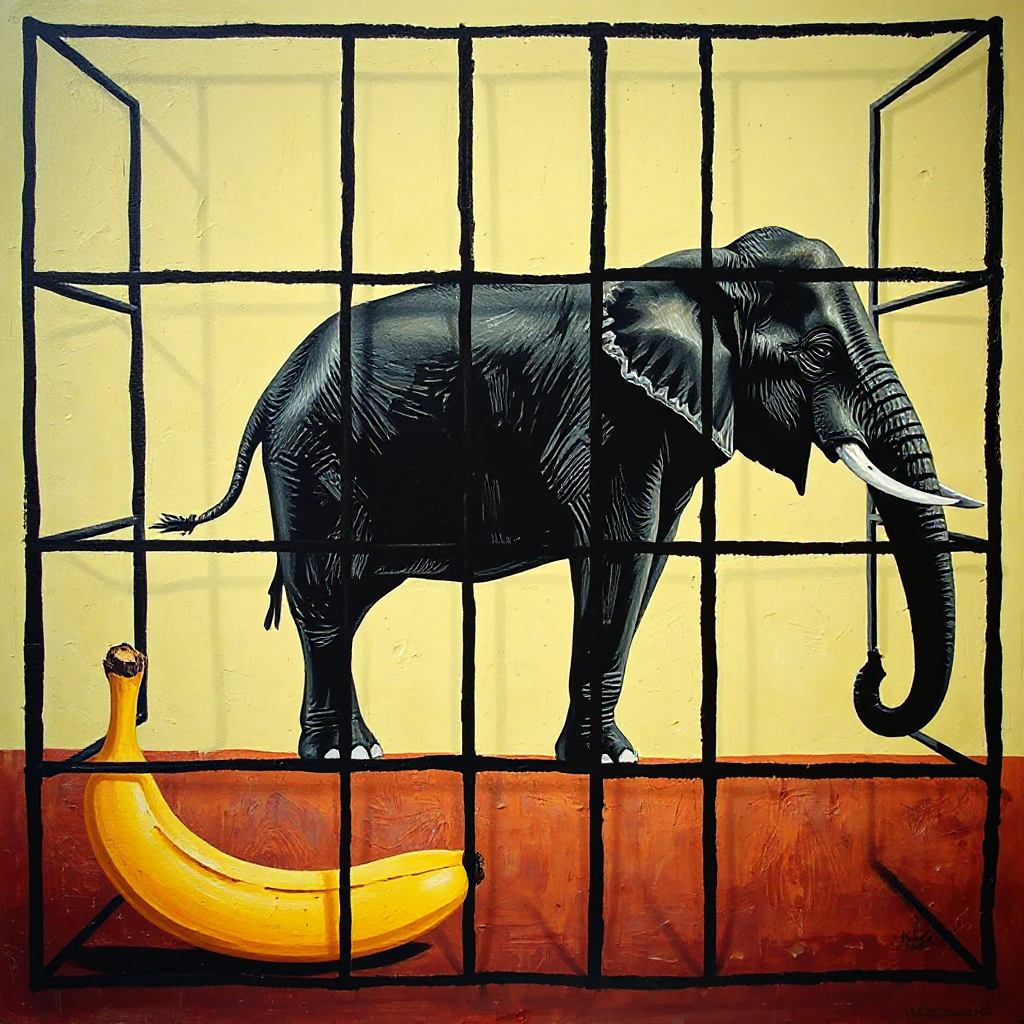}
        \caption{}
    \end{subfigure}
    \hfill
    \begin{subfigure}{0.30\linewidth}
        \centering
        \includegraphics[width=\linewidth]{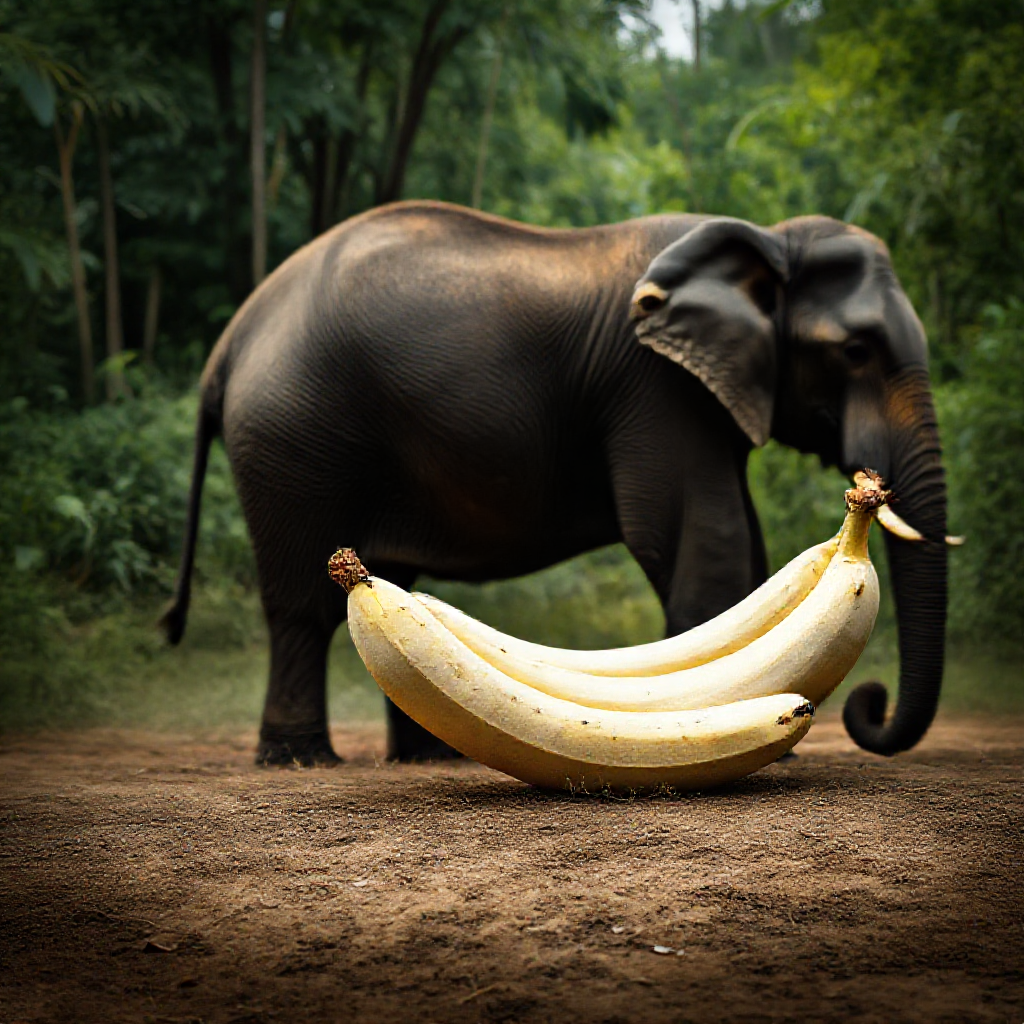}
        \caption{}
    \end{subfigure}

    \caption{``a photo of a white banana and a black elephant''}
    \label{fig: example11}
\end{figure}

\begin{figure}[th]
    \centering
    \begin{subfigure}{0.30\linewidth}
        \centering
        \includegraphics[width=\linewidth]{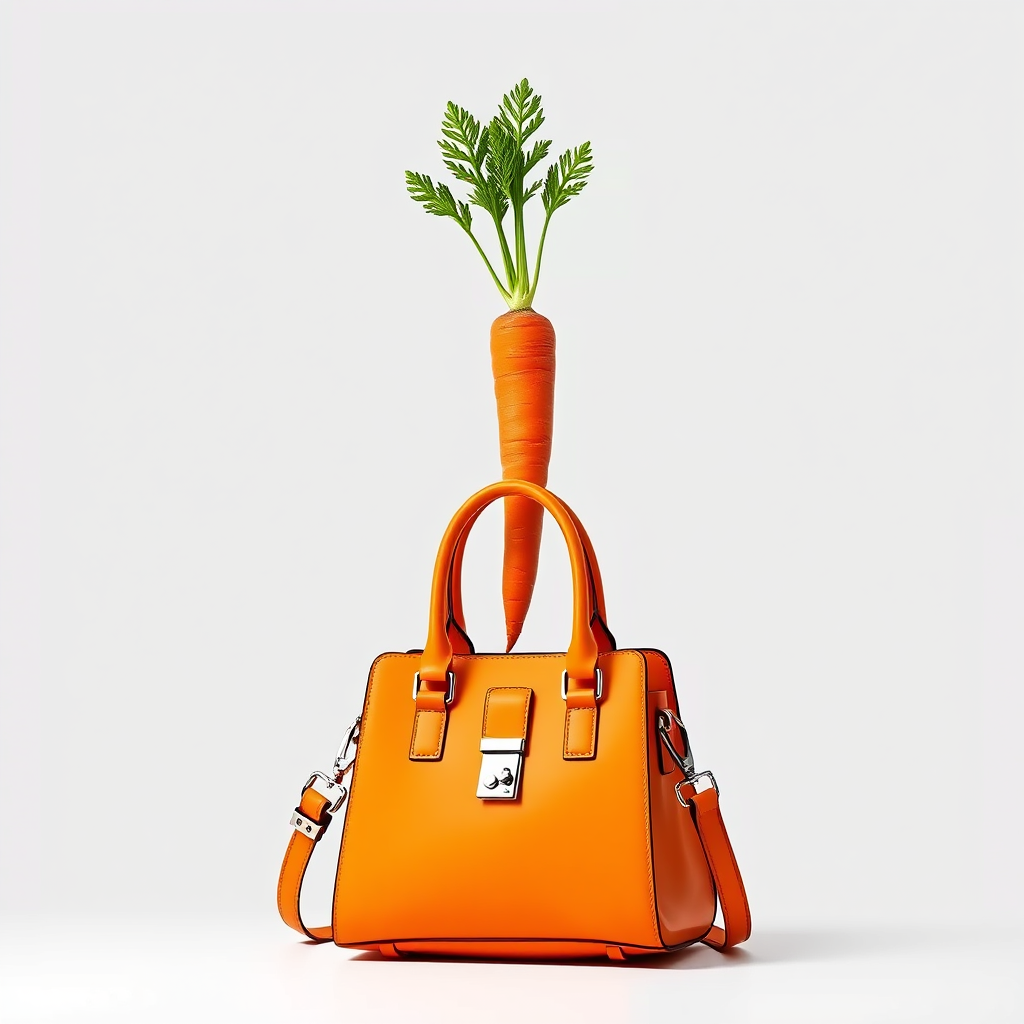}
        \caption{}
    \end{subfigure}
    \hfill
    \begin{subfigure}{0.30\linewidth}
        \centering
        \includegraphics[width=\linewidth]{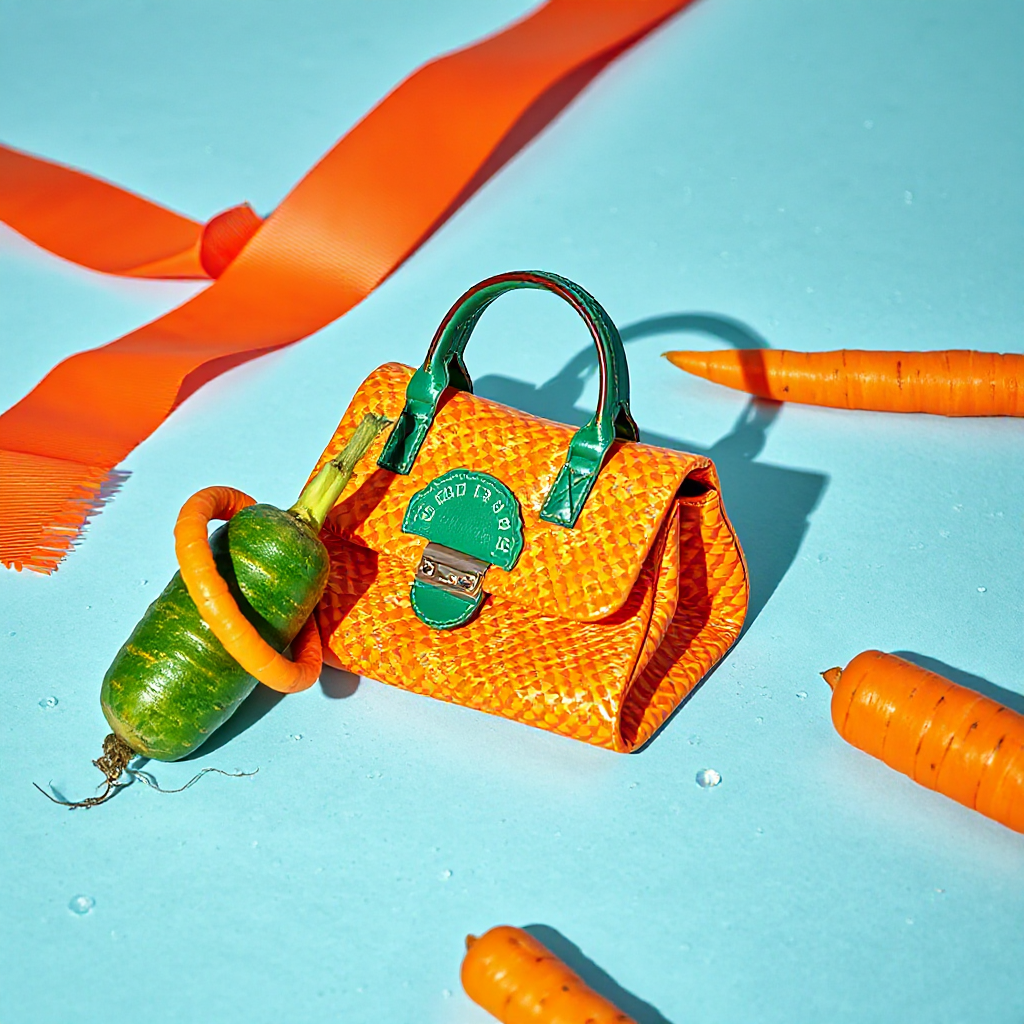}
        \caption{}
    \end{subfigure}
    \hfill
    \begin{subfigure}{0.30\linewidth}
        \centering
        \includegraphics[width=\linewidth]{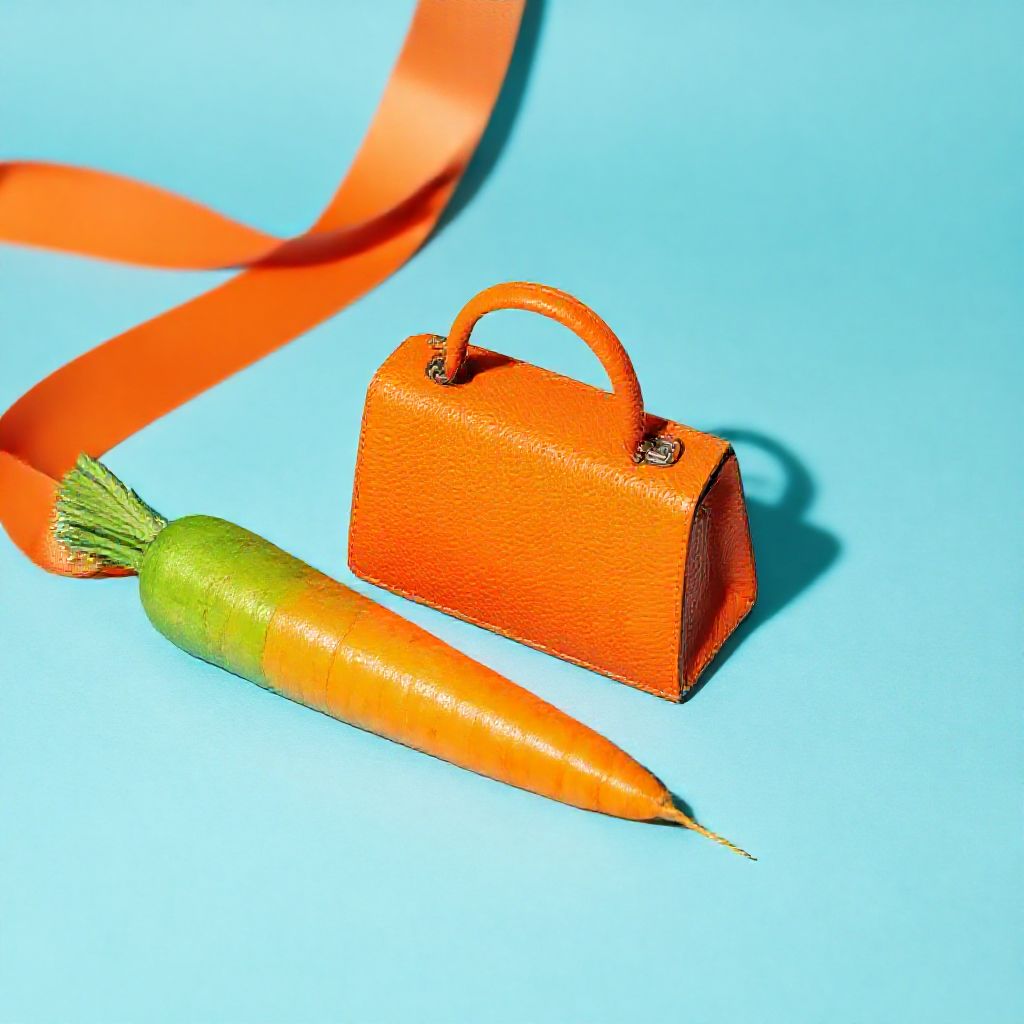}
        \caption{}
    \end{subfigure}

    \caption{``a photo of an orange handbag and a green carrot''}
    \label{fig: example12}
\end{figure}

\begin{figure}[th]
    \centering
    \begin{subfigure}{0.30\linewidth}
        \centering
        \includegraphics[width=\linewidth]{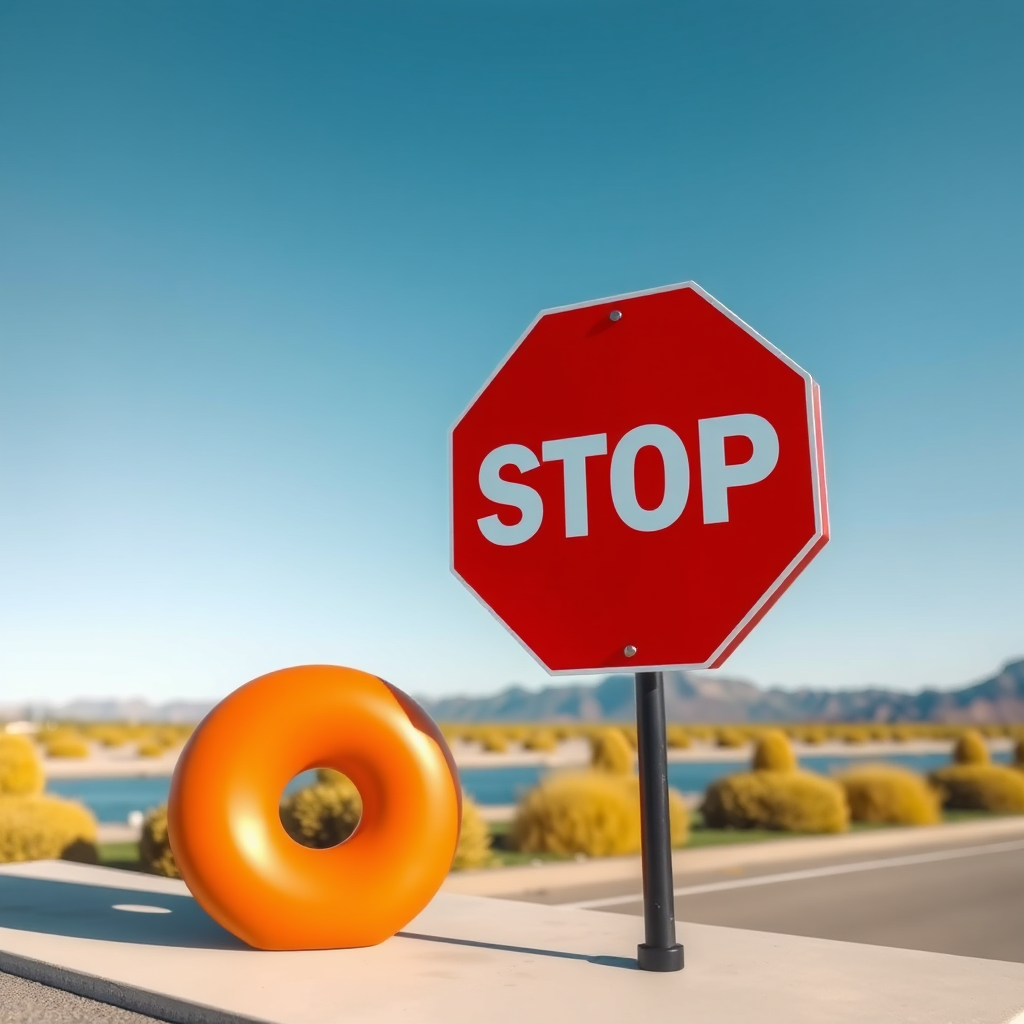}
        \caption{}
    \end{subfigure}
    \hfill
    \begin{subfigure}{0.30\linewidth}
        \centering
        \includegraphics[width=\linewidth]{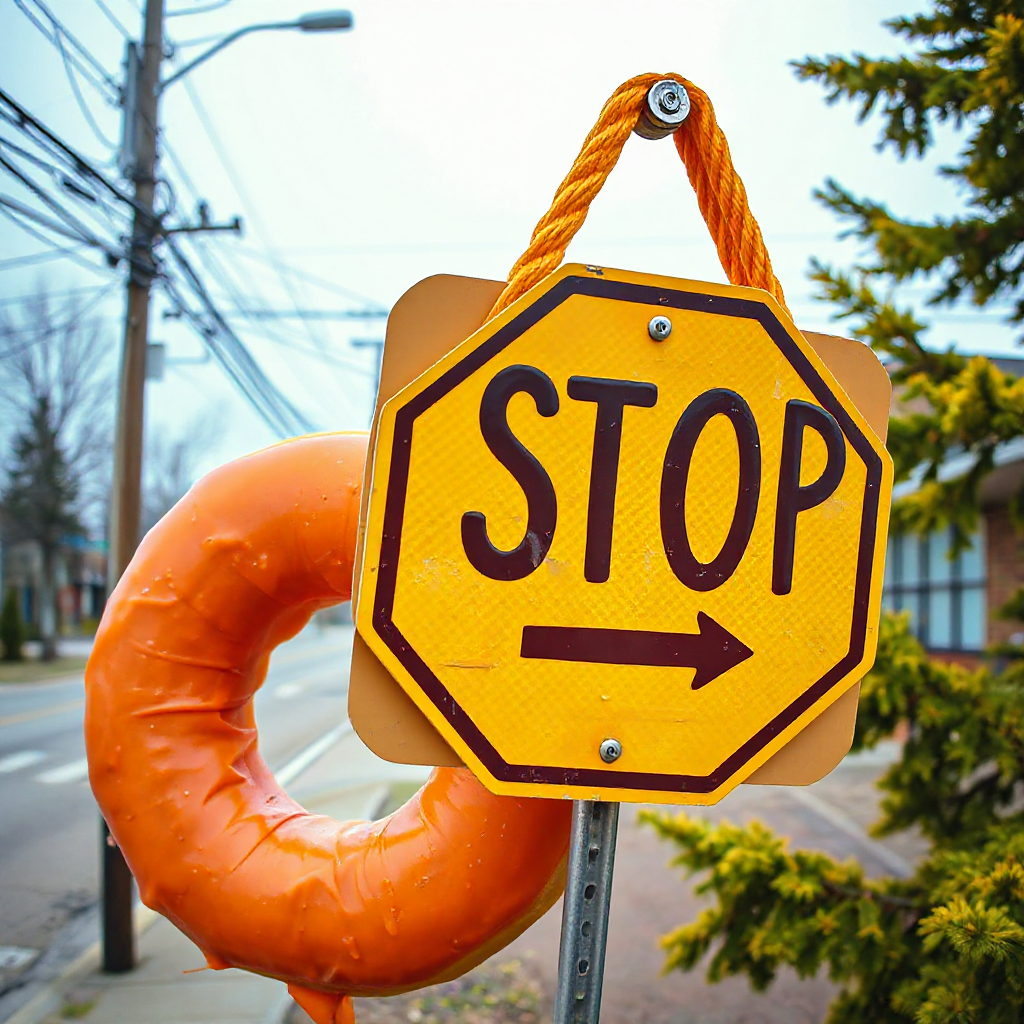}
        \caption{}
    \end{subfigure}
    \hfill
    \begin{subfigure}{0.30\linewidth}
        \centering
        \includegraphics[width=\linewidth]{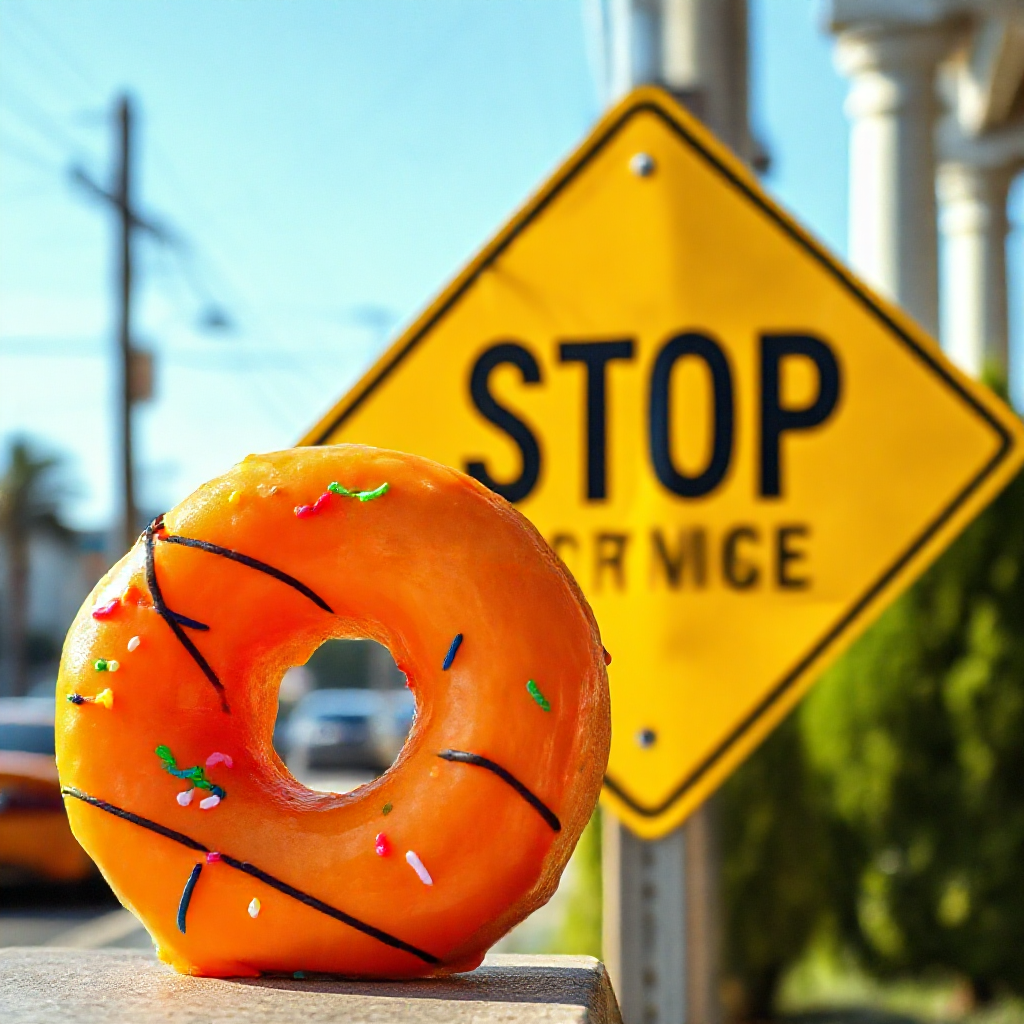}
        \caption{}
    \end{subfigure}

    \caption{``a photo of an orange donut and a yellow stop sign''}
    \label{fig: example13}
\end{figure}

\begin{figure}[th]
    \centering
    \begin{subfigure}{0.30\linewidth}
        \centering
        \includegraphics[width=\linewidth]{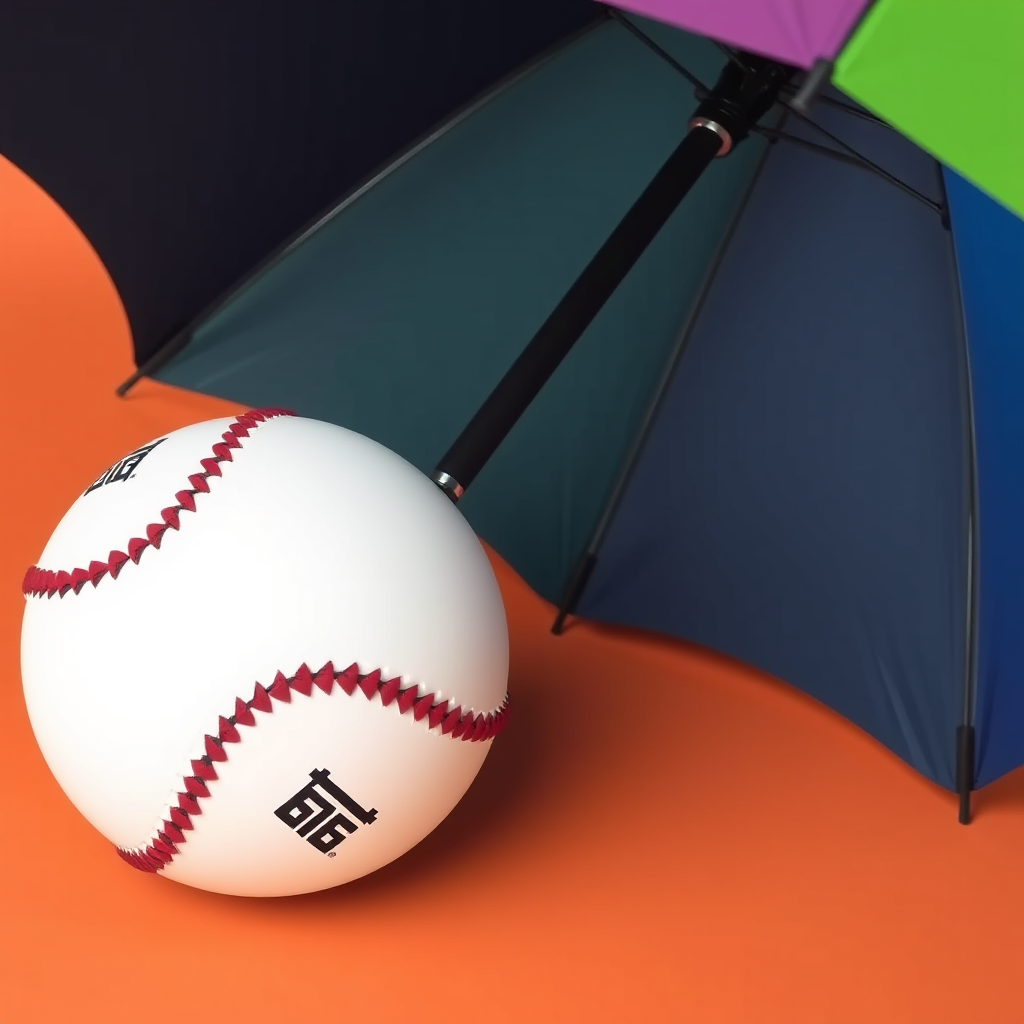}
        \caption{}
    \end{subfigure}
    \hfill
    \begin{subfigure}{0.30\linewidth}
        \centering
        \includegraphics[width=\linewidth]{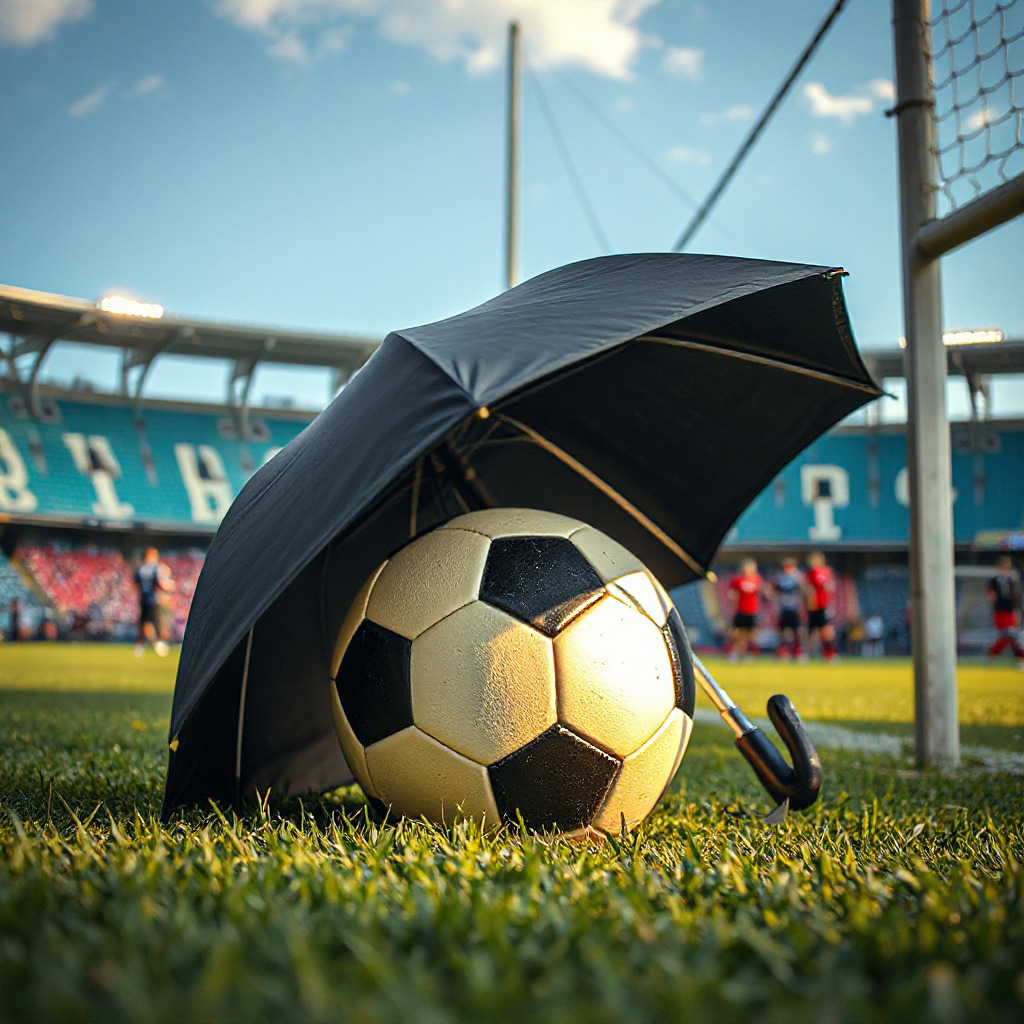}
        \caption{}
    \end{subfigure}
    \hfill
    \begin{subfigure}{0.30\linewidth}
        \centering
        \includegraphics[width=\linewidth]{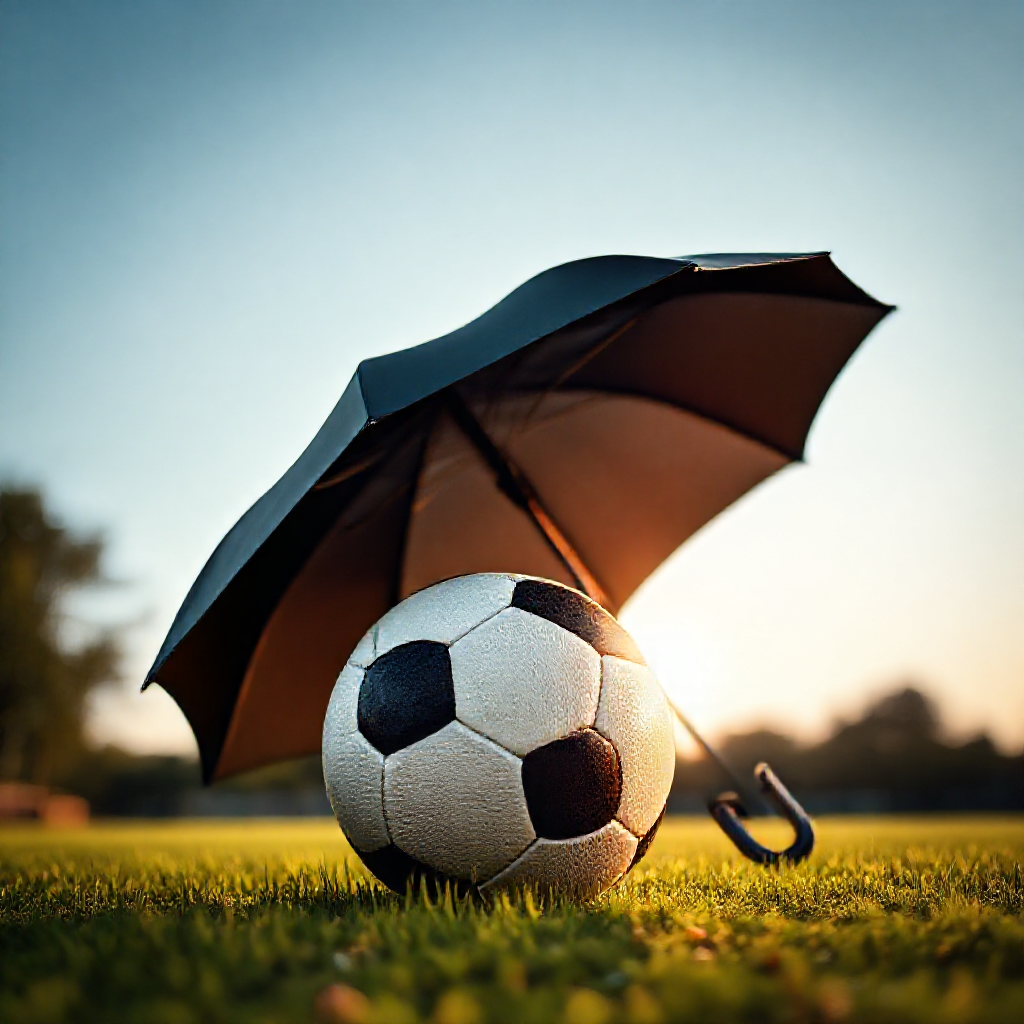}
        \caption{}
    \end{subfigure}

    \caption{``a photo of a sports ball left of an umbrella''}
    \label{fig: example14}
\end{figure}


\end{document}